\documentclass[11pt]{article}
\usepackage{acl}

\usepackage{times}
\usepackage{latexsym}
\usepackage[T1]{fontenc}
\usepackage[utf8]{inputenc}
\usepackage{placeins}
\usepackage{microtype}
\usepackage{inconsolata}
\usepackage{booktabs}
\usepackage{multirow}
\usepackage{amsmath}
\usepackage{amssymb}
\usepackage{graphicx}
\usepackage{caption}
\usepackage{subcaption}
\usepackage{enumitem}
\usepackage{xspace}
\usepackage{url}
\usepackage{xcolor}

\usepackage{comment}
\usepackage{tikz}

\usepackage{algorithm}
\usepackage{algpseudocode}
\newcommand{\stepktomath}{\textsc{Step-KTOder}}
\usepackage[most]{tcolorbox}
\usepackage{xcolor}
\definecolor{prettyblue}{HTML}{2563EB}
\definecolor{prompttitle}{HTML}{2C3E50}    
\definecolor{promptbg}{HTML}{F5F5FA}       

\newtcolorbox{promptbox}[1]{
    enhanced,
    colback=promptbg,
    colframe=prompttitle,
    coltitle=white,
    fonttitle=\bfseries,
    title=#1,
    boxrule=0pt,
    toptitle=2mm, bottomtitle=2mm,
    left=4mm, right=4mm, top=2mm, bottom=2mm,
    fontupper=\ttfamily\footnotesize,
}

\newcommand{\stepkto}{\textsc{Step-KTOder}\xspace}

\usepackage{calc}

\newcommand{\ms}[2]{\makebox[\widthof{$\mathbf{0.000_{\pm0.000}}$}][c]{$#1_{\pm#2}$}}
\newcommand{\msb}[2]{\makebox[\widthof{$\mathbf{0.000_{\pm0.000}}$}][c]{$\mathbf{#1_{\pm#2}}$}}

\usepackage{calc}  

\newcommand{\pv}[1]{\makebox[\widthof{\textbf{0.000}}][c]{#1}}
\newcommand{\pvb}[1]{\makebox[\widthof{\textbf{0.000}}][c]{\textbf{#1}}}

\title{Function-Level Execution Feedback for Code Preference Optimization}

\author{
  \textbf{Idris Nechnech}$^{1}$, 
  \textbf{Sehwan Kim}$^{1}$, 
  \textbf{Jimin Seo}$^{1}$, 
  \textbf{Yeongoon Kim}$^{1}$, \\
  \textbf{Minhae Oh}$^{1}$, 
  \textbf{Sangwoo Hong}$^{2}$, 
  \textbf{Jungwoo Lee}$^{1\dagger}$ \\
  $^{1}$Department of Electrical and Computer Engineering, Seoul National University \\
  $^{2}$Department of Computer Science and Engineering, Konkuk University \\
  \texttt{\{inechnech,junglee\}@snu.ac.kr}
}

\begin{document}
\maketitle
\begingroup\def\thefootnote{$\dagger$}
\footnotetext{Corresponding author.}
\endgroup

\begin{abstract}
Process supervision has improved mathematical reasoning, where intermediate steps are naturally expressed as chains of thought. In code generation, however, process supervision remains underexplored because there is no standard notion of a step. Supervision can target lines, reasoning traces, or program states, making it unclear what to label and optimize. We propose \stepkto, a framework for code preference optimization that defines steps as module-level functions in decomposed multi-function programs and assigns binary correctness labels via automatically generated unit tests. Our method provides a code-specific instantiation of stepwise KTO, combining function-level process supervision with outcome-level feedback on the full program. We evaluate on HumanEval(+), MBPP(+), BigCodeBench, and LiveCodeBench, showing that \stepkto improves over outcome-only KTO and DPO. Further analysis shows that execution-based labels are essential: LLM-as-a-judge annotations systematically over-predict function failures, corrupt positive step labels, and degrade downstream preference optimization. Code is available at: \href{https://github.com/inechnech/STEP-KTODER}
{\textcolor{prettyblue}{\nolinkurl{https://github.com/inechnech/STEP-KTODER}}}.
\end{abstract}

\section{Introduction}

Code generation has become a major application of Large Language Models (LLMs), with recent code-specialized models achieving strong performance on a wide range of programming tasks \citep{guo2024deepseekcoder, hui2024qwen25coder}. These models are typically improved through two stages: supervised fine-tuning (SFT)~\citep{wei2022finetuned} on high-quality code data, followed by an alignment stage that adapts the model toward better outputs \citep{10.5555/3600270.3602281}. Among alignment methods, Direct Preference Optimization (DPO) \citep{rafailov2024dpo} is widely used to learn from paired preference data, whereas Kahneman--Tversky Optimization (KTO) \citep{ethayarajh2024kto} learns from binary desirability labels on individual outputs.

However, applying these techniques to code generation is nontrivial. These methods are typically instantiated at the level of the entire program, treating a full solution as a single unit of feedback. This provides only a coarse training signal: a program can be mostly correct and fail only on a narrow edge case, but outcome-level supervision still marks the entire output as undesirable. Such supervision does not reveal which component is correct and which is responsible for the failure, motivating finer-grained supervision for code generation.

In mathematical reasoning, \emph{process supervision}---providing feedback on intermediate reasoning steps---has proven highly effective \citep{lightman2023lets, luo2024improve, wang2024mathshepherd}, with process reward models (PRMs) underpinning state-of-the-art reasoning systems. Building on this success, \citet{lin2025stepkto} combined stepwise binary feedback with KTO to jointly optimize intermediate and final-answer quality for mathematical problem solving. Transferring this idea to code generation, however, remains an open challenge: code does not naturally decompose into a single standardized sequence of intermediate logical steps, making the very definition of a ``step'' ambiguous.

\begin{figure*}[t]
    \centering
    \includegraphics[width=0.90\textwidth]{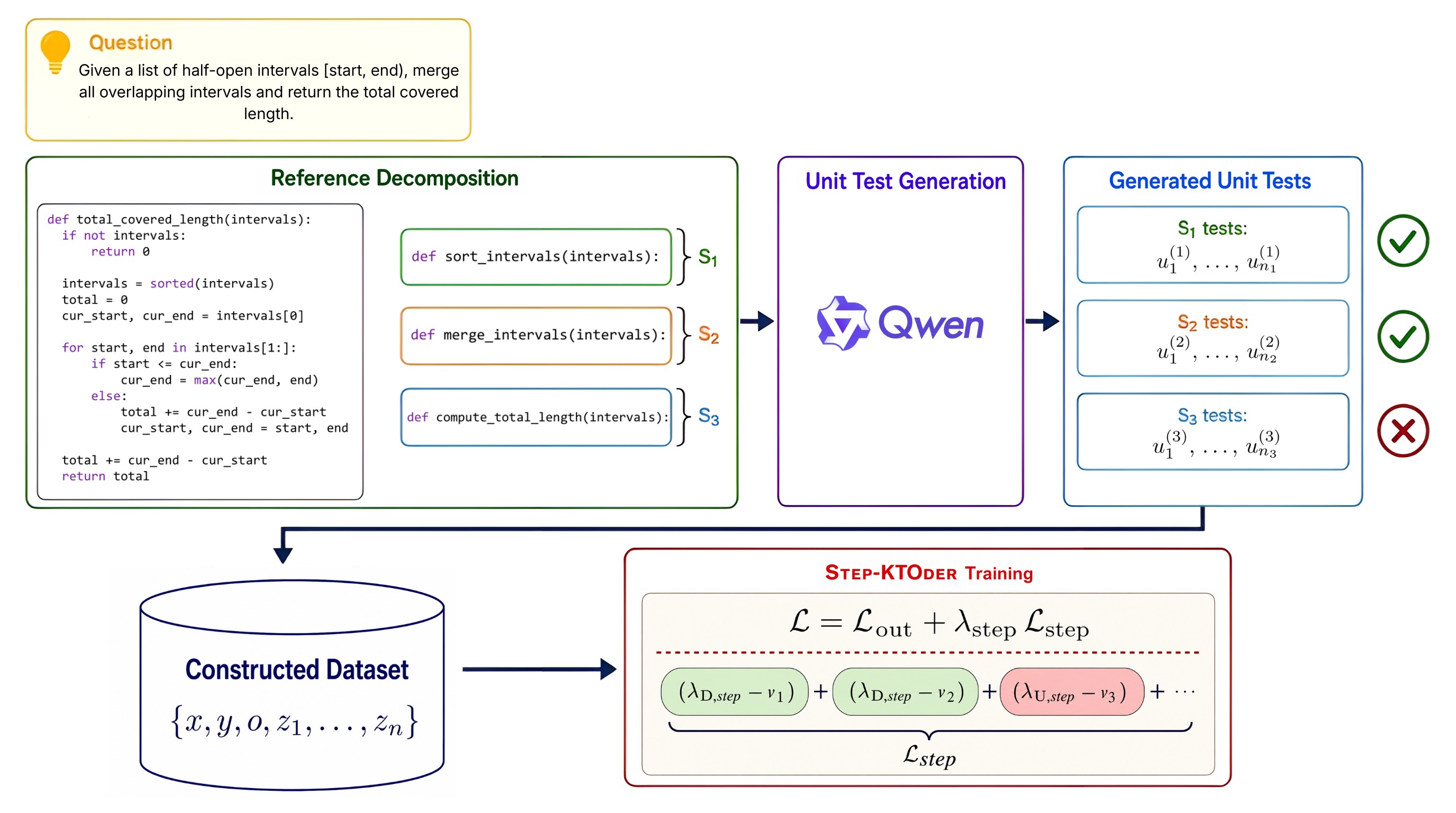}
   \caption{Overview of the \stepkto framework. A reference solution $y^\star$ is decomposed into module-level functions $s_1, \ldots, s_n$, and function-level unit tests are generated for each function. Candidate completions are evaluated against these tests to obtain step labels $z_i$ indicating local correctness, alongside an outcome label $o$ from the dataset-provided test suite. The resulting labeled data is used to train the model with a joint objective that combines outcome-level KTO ($\mathcal{L}_{\mathrm{out}}$) and function-level supervision ($\mathcal{L}_{\mathrm{step}}$), reinforcing locally correct functions while penalizing locally incorrect ones.}
    \label{fig:pipeline}
\end{figure*}

We address this ambiguity by defining steps as module-level functions in decomposed programs, and propose \textbf{\stepkto}, a framework that brings function-level process supervision to code generation. Our key idea is that decomposed programs provide an executable step structure: when a solution is written as multiple functions, each can be tested independently using automatically generated unit tests, yielding binary labels of local correctness. \stepkto uses these labels to extend outcome-level KTO with function-level supervision, guiding training by both global program success and local correctness, as indicated in Figure~\ref{fig:pipeline}.

Beyond the main empirical gains, we investigate two design choices that are important to \stepkto's effectiveness. First, programs that pass dataset-provided tests but contain a locally incorrect function provide a form of supervision that outcome-only training cannot express. We find that preserving these local/global mismatches strengthens the benefit of step-level supervision, whereas masking such mismatches substantially weakens the signal. Second, we assess whether automatically generated unit tests are necessary for reliable step supervision by replacing execution-based step labels with LLM-as-a-judge annotations. Although these annotations show moderate agreement with execution-based labels, they systematically over-predict step failures, and when used for \stepkto training, they degrade performance. Even for already instruction-tuned Qwen2.5-Coder models, \stepkto achieves substantial improvements: it improves over the base model by up to +26.7\% on BigCodeBench Hard \citep{zhuo2025bigcodebench} and +27.0\% on LiveCodeBench \citep{jain2025livecodebench}.

Our contributions are as follows:
\begin{itemize}[leftmargin=*,itemsep=2pt,topsep=2pt]
    \item We propose \stepkto, a code-specific instantiation of stepwise KTO that combines function-level process supervision with outcome-level feedback.
    \item We introduce an automatic data construction pipeline that decomposes solutions into functions, generates function-level unit tests, and derives execution-based step labels for training.
    \item Experiments across widely used benchmarks show that \stepkto improves post-trained code models. We further demonstrate that replacing execution-based step labels with LLM-as-a-judge annotations weakens the step-level signal and degrades performance.
\end{itemize}

\section{Related Work}

\subsection{Process Supervision for Reasoning}

Process supervision has been highly effective for mathematical reasoning, where step-level feedback improves over outcome-only supervision \citep{cobbe2021trainingverifierssolvemath, uesato2022solvingmathwordproblems, lightman2023lets}. Subsequent work has scaled this supervision: \citet{wang2024mathshepherd} introduced automatic step-level annotation via Monte Carlo estimation, and \citet{luo2024improve} proposed a divide-and-conquer Monte Carlo Tree Search framework for large-scale process supervision data. \citet{lin2025stepkto} introduced Step-KTO, which combines KTO with PRM-labeled stepwise feedback to jointly optimize intermediate reasoning quality and final response correctness. Our work builds on this idea but reformulates it for code: instead of using PRM-labeled reasoning steps, we define steps as concrete, independently testable functions and supervise them directly through execution-based unit test feedback.
This perspective is consistent with recent work on code PRMs showing that execution feedback improves the reliability of process supervision for code~\citep{li2025codeprm}. In contrast to PRM training, we study how execution-based process labels can be incorporated directly into offline preference optimization.

\subsection{Preference Optimization for Code}

Preference optimization has become a standard post-training technique for code models. DPO \citep{rafailov2024dpo} learns from paired preferences, while KTO \citep{ethayarajh2024kto} removes the requirement for paired data using binary desirability labels grounded in prospect theory. Recent work has explored localized and fine-grained feedback for code: \citet{zhang2025focuseddpo} concentrated the DPO loss on error-prone token spans, while \citet{wu-etal-2025-teaching} introduced Target-DPO, a focal preference alignment framework that localizes preference updates to targeted code regions. \citet{dai2024process} trained a PRM that provides dense line-level feedback during code generation via reinforcement learning. \citet{zhang2025codedpo} introduced CodeDPO, which constructs self-generated preference pairs to align code models for both correctness and efficiency.

These approaches target different granularities---tokens, code blocks, lines, or whole programs---but to our knowledge, none of these approaches defines steps as semantically meaningful, independently testable functions and integrates them into a binary preference optimization objective.

\subsection{Solution Decomposition and Automated Testing}

Fill-in-the-middle objectives \citep{bavarian2022efficienttraininglanguagemodels} train models to complete code given surrounding context, encouraging modular structure. \citet{ren-etal-2025-alignment} extend this idea to alignment, splitting code into AST-based blocks to construct more diverse preference pairs for DPO.
Prompting-based approaches have also emphasized modular decomposition for code generation: \citet{pan2025modularization} break complex programming problems into smaller reasoning modules through hierarchical prompting. \citet{lin2025solvers} proposed generating both code and unit tests simultaneously, then using self-validation to enhance generation quality. Automated unit test generation has matured significantly \citep{chen2023codet, ma-etal-2025-dynamic}; our pipeline leverages this capability in a function-level setting, generating tests for individual functions and using execution outcomes as binary step labels for training.

\subsection{Execution-Grounded Inference-Time Methods}
Recent execution-grounded methods instead improve code at inference time. S$^\ast$ combines parallel sampling with iterative debugging and selects among candidates using adaptively synthesized inputs and their execution results~\citep{li-etal-2025-test}, while ORPS explores a tree of reasoning and code trajectories guided by execution outcomes and self-critique~\citep{yu2025reasoning}. Unlike these per-query search and selection methods, \stepkto uses function-level execution during offline data construction to produce labels that update the model policy.
\section{Method}
\label{sec:method}

\subsection{Problem Formulation}
\label{sec:problem-formulation}
Given a prompt $x$ and a reference solution $y^\star$, we first rewrite $y^\star$ into a decomposed multi-function program
\[
\tilde{y} = (s_1, s_2, \dots, s_n),
\]
using a strong code language model, where each $s_i$ denotes one function in the decomposed program and $n$ is the total number of such functions. We treat each function $s_i$ as a \emph{step}.

Training samples are then generated from this decomposition: the target model fills the function skeleton defined by $\tilde{y}$ to produce a candidate solution $y$. Each training sample receives two forms of binary supervision:
\begin{itemize}[leftmargin=*,itemsep=2pt,topsep=2pt]
    \item An \textbf{outcome label} $o \in \{0,1\}$, indicating whether $y$ passes the dataset-provided test suite for $x$.
    \item A sequence of \textbf{step labels} $\mathbf{z}=(z_1,\dots,z_n)$, where each $z_i \in \{0,1,\varnothing\}$ indicates whether the implementation of $s_i$ in $y$ passes its function-level unit tests. A label of $\varnothing$ indicates that no valid unit test is available for that step.
\end{itemize}

This formulation distinguishes \emph{local correctness} from \emph{global correctness}, providing finer-grained supervision than outcome-only evaluation. Rather than treating partially correct programs as uniformly desirable or undesirable, it enables the objective to reinforce correct functions while penalizing those responsible for failure.

\subsection{KTO Background}
\label{sec:kto_background}

We build on Kahneman--Tversky Optimization (KTO)~\citep{ethayarajh2024kto}, which aligns a policy $\pi_\theta$ from binary feedback using a Kahneman--Tversky-inspired value function over the log-ratio between the policy and a frozen reference $\pi_{\mathrm{ref}}$:
\[
r_\theta(x, y)
=
\log \frac{\pi_\theta(y \mid x)}{\pi_{\mathrm{ref}}(y \mid x)} .
\]

The outcome-level reference point is
\[
z_0^{\mathrm{out}}
=
\mathrm{KL}\!\left(
\pi_\theta(y' \mid x)
\;\middle\|\;
\pi_{\mathrm{ref}}(y' \mid x)
\right),
\]
where $y'$ denotes an output sequence used to estimate the divergence between the policy and reference. Given a binary desirability label $o \in \{0,1\}$, the outcome-level value function is
\[
v_{\mathrm{out}}=
\begin{cases}
\lambda_D \, \sigma\!\bigl(\beta_{\mathrm{out}} (r_\theta-z_0^{\mathrm{out}})\bigr), & o = 1, \\[2pt]
\lambda_U \, \sigma\!\bigl(\beta_{\mathrm{out}} (z_0^{\mathrm{out}}-r_\theta)\bigr), & o = 0,
\end{cases}
\]
where $r_\theta \equiv r_\theta(x,y)$ and  $\sigma(\cdot)$ denotes the sigmoid function. $\beta_{\mathrm{out}} > 0$ controls the sensitivity of the outcome-level value function, and $\lambda_D$ and $\lambda_U$ weight desirable and undesirable samples, respectively. The outcome-level KTO loss is then
\[
\mathcal{L}_{\mathrm{out}}(\pi_\theta,\pi_{\mathrm{ref}})
=
\mathbb{E}_{(x,y,o)\sim\mathcal{D}}
\bigl[
\lambda_o - v_{\mathrm{out}}
\bigr],
\]
where $\lambda_o=\lambda_D$ if $o=1$ and $\lambda_o=\lambda_U$ if $o=0$.

\subsection{Step-KTO Objective}
\label{sec:stepkto_objective}

Following \citet{lin2025stepkto}, we instantiate stepwise KTO over the function-level steps defined in Section~\ref{sec:problem-formulation} and introduce reliability masking for functions without validated local tests. For each step $s_i$, we compute a step-local log-ratio over the tokens of that function:
\[
r_i
=
\sum_{t \in s_i}
\log
\frac{\pi_\theta(y_t \mid x, y_{<t})}
{\pi_{\mathrm{ref}}(y_t \mid x, y_{<t})}.
\]
Each step has a label $z_i$ as defined above and a mask $m_i \in \{0,1\}$; only steps with $m_i=1$ and $z_i\in\{0,1\}$ contribute to the step-level loss.

Similarly, we define a step-level reference point
\[
z_0^{\mathrm{step}}
=
\mathrm{KL}\!\left(
\pi_\theta(y'_i \mid x, s_{<i})
\;\middle\|\;
\pi_{\mathrm{ref}}(y'_i \mid x, s_{<i})
\right),
\]
where $y'_i$ denotes the subsequence of generated tokens belonging to the $i$-th function. The step-level value function is
\[
v_i =
\begin{cases}
\lambda_{D,\mathrm{step}} \, \sigma\!\bigl(\beta_{\mathrm{step}} (r_i-z_0^{\mathrm{step}})\bigr), & z_i = 1, \\[2pt]
\lambda_{U,\mathrm{step}} \, \sigma\!\bigl(\beta_{\mathrm{step}} (z_0^{\mathrm{step}}-r_i)\bigr), & z_i = 0.
\end{cases}
\]
Here, $\beta_{\mathrm{step}}$, $\lambda_{D,\mathrm{step}}$, and $\lambda_{U,\mathrm{step}}$ are the step-level counterparts of the outcome-level KTO parameters.

Let $\mathcal{M}_\tau=\{i:m_i=1,\ z_i\in\{0,1\}\}$ denote the supervised steps for sample $\tau$, and let $M=|\mathcal{M}_\tau|$. Let $\lambda_{z_i,\mathrm{step}}$ denote the step-level target coefficient, with $\lambda_{z_i,\mathrm{step}}=\lambda_{D,\mathrm{step}}$ when $z_i=1$ and $\lambda_{z_i,\mathrm{step}}=\lambda_{U,\mathrm{step}}$ when $z_i=0$. The masked step-level loss is
\[
\mathcal{L}_{\mathrm{step}}
=
\mathbb{E}_{\tau\sim\mathcal{D}}
\left[
\frac{1}{n}
\sum_{i\in\mathcal{M}_\tau}
\left(
\lambda_{z_i,\mathrm{step}}-v_i
\right)
\right],
\]
where the sum is zero when $M=0$, so samples without supervised steps fall back to outcome-only KTO.

\begin{figure*}[t]
    \centering
    \includegraphics[width=0.90\textwidth]{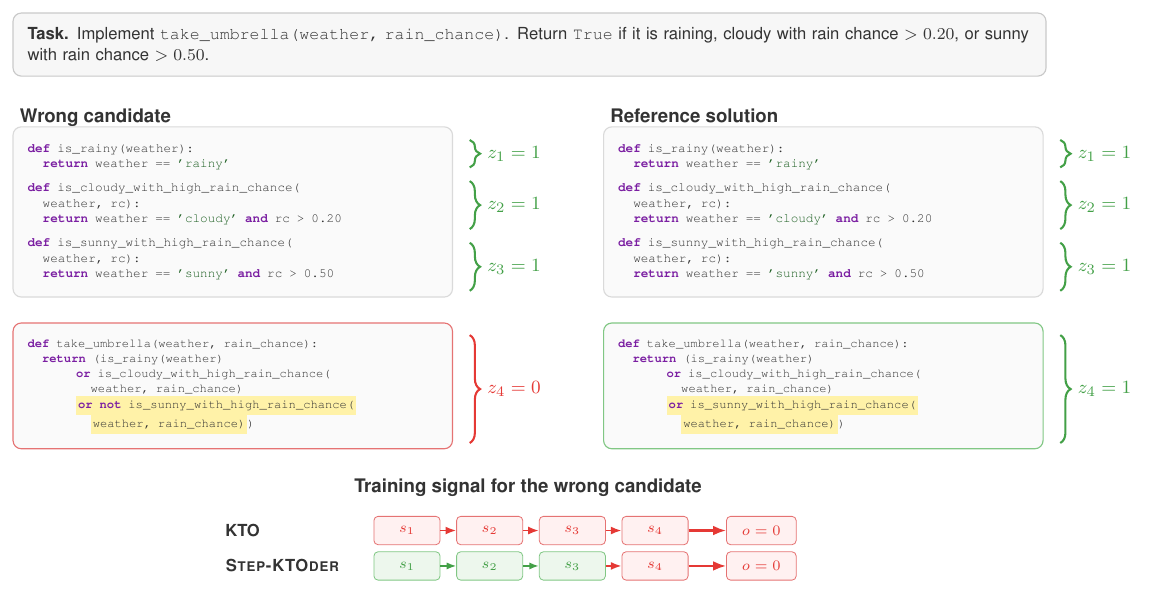}
    \caption{Training signal comparison for a failing candidate. The wrong candidate differs from the decomposed reference solution only in the highlighted line, yielding locally correct functions $z_1,z_2,z_3=1$, a faulty final function $z_4=0$, and outcome label $o=0$. Outcome-only KTO penalizes all function spans, whereas \stepkto preserves the locally correct functions and penalizes only the faulty one.}
    \label{fig:function_supervision_example}
\end{figure*}

Combining with the outcome-level loss yields the final \stepkto objective:
\[
\mathcal{L}_{\text{\stepktomath}}
=
\mathcal{L}_{\mathrm{out}}
+
\lambda_{\mathrm{step}}\,\mathcal{L}_{\mathrm{step}} .
\]
By jointly optimizing outcome-level and function-level feedback, \stepkto reinforces locally correct functions and penalizes locally incorrect ones, providing a more localized learning signal than outcome-level supervision alone. Figure~\ref{fig:function_supervision_example} illustrates this difference on a failing candidate: outcome-only KTO treats the full program as undesirable, whereas \stepkto assigns negative step-level signal only to the faulty function.

\subsection{Data Construction Pipeline}
\label{sec:pipeline}

Our pipeline consists of four stages, illustrated in Figure~\ref{fig:pipeline}.

\paragraph{(1) Reference decomposition.}
We use a strong instruction-tuned code model to rewrite each ground-truth solution into an interface-preserving multi-function program, retaining only rewrites that pass the original test suite.

\paragraph{(2) Function-level unit test generation.}
For each decomposed function, a strong code model generates several targeted unit tests for its interface and edge cases. We retain only tests that pass on the reference implementation, invoke the target function, and contain nontrivial assertions whose pass condition depends on the function's output. Functions without valid tests receive a null step label ($z_i = \varnothing$) and are masked from the step loss. The prompts are provided in Appendix~\ref{app:prompts}.

\paragraph{(3) On-policy candidate generation.}
From each decomposition, we construct a \emph{skeleton} by replacing function bodies with \texttt{pass} while preserving signatures, docstrings, imports, and module-level context. Then, for each skeleton we sample $k=8$ candidate completions from the model being post-trained, using temperature $T=0.4$ and top-$p$ sampling with $p=0.95$; this keeps the preference data close to the model's own generation distribution.

\begin{figure}[t]
    \centering
    \includegraphics[width=\columnwidth]{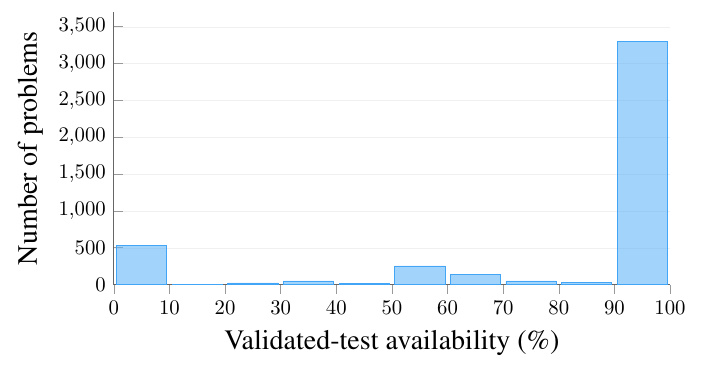}
    \caption{Distribution of per-problem validated-test availability. Most problems achieve near-full test availability.}
    \label{fig:ut_coverage}
\end{figure}

\paragraph{(4) Labeling and dataset construction.}
Each candidate is evaluated against the dataset-provided test suite to obtain an outcome label $o$. Step labels $z_i$ are obtained by executing each candidate function against its generated unit tests. By default, we use \emph{local} step labels: each function is tested after being inserted into the decomposed reference program, with all other functions kept fixed to their reference implementations. This isolates function-level correctness from downstream composition effects. Across datasets, 83.2\% of testable decomposed functions retain at least one
validated test. We refer to this quantity as \emph{validated-test availability}; the corresponding per-problem distribution is shown in Figure~\ref{fig:ut_coverage}. For each task, we retain up to one passing and one failing candidate, preferring candidates with richer step-level supervision. Rows with fewer than two supervised steps fall back to outcome-only KTO, avoiding single-step supervision that largely duplicates the outcome-level signal.

\subsection{Conflict-Preserving Step Labels}
\label{sec:conflicts}

A key design question is how to handle examples where outcome-level and step-level supervision disagree. Two conflict patterns arise: \textbf{passing programs with negative step labels}, where the full program passes dataset tests but at least one supervised function fails its function-level tests; and \textbf{failing programs with all-positive step labels}, where the full program fails dataset tests even though all supervised functions pass their unit tests.
These examples reflect a mismatch between local and end-to-end correctness. Rather than removing them, we compare masked and conflict-preserving variants to test whether such cases provide useful training signal beyond outcome-only supervision and ordinary partial-correctness patterns.

\section{Experimental Setup}
\label{sec:setup}

\subsection{Training Configuration}

All target models are instruction-tuned code models that have already undergone substantial post-training. We fine-tune each model for 1 epoch using LoRA~\citep{hu2022lora} with global batch size 16. KTO and \stepkto use a learning rate of $1\times10^{-6}$, while DPO uses $5\times10^{-7}$. We set $\lambda_{\mathrm{step}}=1.0$ by default. Full training details are provided in Appendix~\ref{app:training_details}.

\subsection{Datasets}
\label{sec:datasets}

Training data is drawn from the training splits of TACO~\citep{li2023taco} and APPS~\citep{hendrycks2021measuring}. We construct training samples using the pipeline described in Section~\ref{sec:method}: reference solutions are decomposed with Qwen2.5-Coder-32B-Instruct~\citep{hui2024qwen25coder}, and each target model generates on-policy candidate completions from the resulting function skeletons. For each problem, we retain up to one passing and one failing candidate using a ranking procedure that prioritizes positives whose supervised functions all pass their unit tests and negatives with both passing and failing function-level labels. Such negatives localize which components remain correct and which fail (Appendix~\ref{app:posneg}).

Table~\ref{tab:dataset_stats} summarizes the resulting training sets. For same-family training with Qwen2.5-Coder-1.5B-Instruct and Qwen2.5-Coder-3B-Instruct, we include decomposed reference solutions as additional positive anchors. For cross-family training with DeepSeek-Coder-6.7B-Instruct, we use only on-policy generated candidates, since adding off-policy samples degrades performance (Appendix~\ref{app:gt_ablation}). Rows with fewer than two supervised steps fall back to outcome-only KTO. As a result, 82\% of Qwen2.5-Coder-1.5B-Instruct rows, 83\% of Qwen2.5-Coder-3B-Instruct rows, and 78\% of DeepSeek-Coder-6.7B-Instruct rows carry active step-level supervision. We justify this threshold in Section~\ref{sec:single_step}.

\subsection{Evaluation Benchmarks and Baselines}
\begin{table}[t]
\centering
\scriptsize
\setlength{\tabcolsep}{3pt}
\resizebox{\columnwidth}{!}{%
\begin{tabular}{lrrr}
\toprule
& Qwen-1.5B & Qwen-3B & DeepSeek \\
\midrule
\multicolumn{4}{l}{\textit{Dataset composition}} \\
\quad Total rows & 11{,}397 & 11{,}440 & 6{,}573 \\
\quad GT rows & 5{,}018 & 5{,}018 & 0 \\
\quad Generated rows & 6{,}379 & 6{,}422 & 6{,}573 \\
\midrule
\multicolumn{4}{l}{\textit{Step supervision}} \\
\quad Step-supervised & 9{,}376 & 9{,}455 & 5{,}157 \\
\quad Fallback KTO & 2{,}021 & 1{,}985 & 1{,}416 \\
\quad Avg.\ steps / active row & 2.66 & 2.66 & 2.68 \\
\midrule
\multicolumn{4}{l}{\textit{Conflict analysis}} \\
\quad Conflicts ($o{=}1,z_i{=}0$) & 1{,}194 & 1{,}397 & 1{,}428 \\
\midrule
\multicolumn{4}{l}{\textit{Structure quality}} \\
\quad AST alignment & 84.2\% & 84.7\% & 80.3\% \\
\bottomrule
\end{tabular}%
}
\caption{Training dataset statistics. Step-supervised rows contain at least two supervised function-level labels and activate the step-level loss; fallback KTO rows use outcome-only KTO. Avg.\ steps / active row is computed over step-supervised rows. AST alignment measures exact agreement with the decomposition skeleton.}
\label{tab:dataset_stats}
\end{table}

We evaluate on seven code generation benchmarks: HumanEval~\citep{chen2021evaluatinglargelanguagemodels} and HumanEval+~\citep{liu2023evalplus}, MBPP~\citep{austin2021programsynthesislargelanguage} and MBPP+~\citep{liu2023evalplus}, BigCodeBench Full and Hard~\citep{zhuo2025bigcodebench}, and LiveCodeBench~\citep{jain2025livecodebench} release \texttt{v4\_v5}. All evaluations use greedy decoding with vLLM~\citep{kwon2023efficient}.

We compare \stepkto against four baselines: the original instruction-tuned model, DPO~\citep{rafailov2024dpo}, KTO~\citep{ethayarajh2024kto}, and Target-DPO~\citep{wu-etal-2025-teaching}. Baseline construction details are provided in Appendix~\ref{app:baseline_details}.

\section{Results}
\label{sec:results}

\subsection{Main Results}

\begin{table*}[t]
\centering
\small
\begin{tabular*}{\textwidth}{@{\extracolsep{\fill}}lccccccc}
\toprule
Method
& \multicolumn{2}{c}{HumanEval}
& \multicolumn{2}{c}{MBPP}
& \multicolumn{2}{c}{BigCodeBench}
& LiveCodeBench \\
\cmidrule(lr){2-3}
\cmidrule(lr){4-5}
\cmidrule(lr){6-7}
& Base & Plus & Base & Plus & Full & Hard &  \\
\midrule
\textbf{Qwen2.5-Coder-1.5B-Instruct}
& 0.689 & 0.604 & \textbf{0.664} & 0.556 & 0.240 & 0.054 & 0.052 \\
\quad w/ DPO
& 0.695 & 0.610 & \textbf{0.664} & \textbf{0.563} & 0.245 & 0.054 & 0.056 \\
\quad w/ Target-DPO
& 0.683 & \textbf{0.628} & \textbf{0.664} & 0.558 & 0.245 & \textbf{0.068} & 0.019 \\
\quad w/ KTO
& \textbf{0.701} & 0.610 & 0.648 & 0.542 & 0.245 & 0.061 & 0.052 \\
\quad w/ \stepkto
& \textbf{0.701} & 0.616 & 0.656 & 0.548 & \textbf{0.248} & \textbf{0.068} & \textbf{0.056} \\
\quad \emph{Relative improvement over KTO}
& \emph{0.0\%} & \emph{+1.0\%} & \emph{+1.2\%} & \emph{+1.1\%} & \emph{+1.2\%} & \emph{+11.5\%} & \emph{+7.7\%} \\
\midrule
\textbf{Qwen2.5-Coder-3B-Instruct}
& 0.860 & 0.799 & 0.741 & 0.619 & 0.356 & 0.101 & 0.100 \\
\quad w/ DPO
& 0.854 & 0.811 & 0.741 & \textbf{0.630} & 0.360 & 0.115 & 0.112 \\
\quad w/ Target-DPO
& 0.854 & 0.805 & \textbf{0.746} & \textbf{0.630} & 0.351 & 0.108 & 0.112 \\
\quad w/ KTO
& \textbf{0.878} & \textbf{0.835} & 0.741 & 0.624 & \textbf{0.367} & 0.115 & 0.116 \\
\quad w/ \stepkto
& \textbf{0.878} & \textbf{0.835} & \textbf{0.746} & \textbf{0.630} & \textbf{0.367} & \textbf{0.128} & \textbf{0.127} \\
\quad \emph{Relative improvement over KTO}
& \emph{0.0\%} & \emph{0.0\%} & \emph{+0.7\%} & \emph{+1.0\%} & \emph{0.0\%} & \emph{+11.3\%} & \emph{+9.5\%} \\
\midrule
\textbf{DeepSeek-Coder-6.7B-Instruct}
& \textbf{0.799} & 0.713 & 0.746 & 0.640 & 0.342 & 0.108 & 0.127 \\
\quad w/ DPO
& \textbf{0.799} & 0.726 & 0.749 & 0.643 & \textbf{0.347} & 0.115 & 0.127 \\
\quad w/ Target-DPO
& \textbf{0.799} & \textbf{0.738} & 0.725 & 0.624 & 0.346 & 0.108 & 0.112 \\
\quad w/ KTO
& 0.793 & 0.726 & 0.749 & 0.640 & \textbf{0.347} & 0.122 & 0.127 \\
\quad w/ \stepkto
& \textbf{0.799} & 0.732 & \textbf{0.754} & \textbf{0.648} & \textbf{0.347} & \textbf{0.128} & \textbf{0.131} \\
\quad \emph{Relative improvement over KTO}
& \emph{+0.8\%} & \emph{+0.8\%} & \emph{+0.7\%} & \emph{+1.2\%} & \emph{0.0\%} & \emph{+4.9\%} & \emph{+3.2\%} \\
\bottomrule
\end{tabular*}
\caption{Pass rate across model families and scales. Relative improvements are computed over outcome-only KTO within each model; gains are largest on harder benchmarks. The best results are highlighted in bold.}
\label{tab:main_results}
\end{table*}

The experimental results are reported in Table~\ref{tab:main_results}. Since all target models are already instruction-tuned code models, easier benchmarks such as HumanEval and MBPP leave limited headroom, and the clearest gains appear on harder benchmarks. On Qwen2.5-Coder-1.5B-Instruct, \stepkto improves over KTO by +11.5\% on BigCodeBench Hard and +7.7\% on LiveCodeBench. On Qwen2.5-Coder-3B-Instruct, the gains are similarly concentrated on the hardest benchmarks, with +11.3\% on BigCodeBench Hard and +9.5\% on LiveCodeBench, while matching or slightly improving KTO elsewhere. DeepSeek-Coder-6.7B-Instruct shows the same qualitative pattern: BigCodeBench Hard improves monotonically from DPO to KTO to \stepkto, with \stepkto achieving a +4.9\% gain over KTO, and LiveCodeBench improving by +3.2\%. On the easier benchmarks, where headroom is limited, \stepkto remains competitive, obtaining the best or tied-best result in most cases, with Target-DPO leading slightly on HumanEval+ for DeepSeek-Coder-6.7B-Instruct. The gains persist across seeds: \stepkto exceeds KTO on both BigCodeBench Hard and LiveCodeBench across three seeds (Appendix~\ref{app:multiseed}).

Both Target-DPO and \stepkto target localized code failures, but they optimize different supervision signals: Target-DPO derives preference pairs from debugging traces and localizes loss to changed token regions, whereas \stepkto labels functions directly via unit tests. \stepkto is consistently stronger on the harder benchmarks. For Qwen2.5-Coder-1.5B-Instruct, Target-DPO is competitive on easier benchmarks and ties on BigCodeBench Hard, but \stepkto is stronger on LiveCodeBench. For Qwen2.5-Coder-3B-Instruct and DeepSeek-Coder-6.7B-Instruct, \stepkto improves over Target-DPO on both BigCodeBench Hard (+18.5\% for both) and LiveCodeBench (+13.4\% and +17.0\%, respectively). On easier benchmarks, \stepkto ties or improves over Target-DPO in most cases. Notably, \stepkto uses at most 11{,}440 training samples, compared with Target-DPO's 59{,}000 preference pairs.

\subsection{Ablations}

Table~\ref{tab:ablations} reports three core ablations on Qwen2.5-Coder-3B-Instruct. \textit{First}, conflict preservation is essential: masking conflicts reduces \stepkto to KTO-level performance on BigCodeBench Hard and decreases on LiveCodeBench as well. This shows that apparent disagreements between outcome-level and function-level labels can provide useful training signal rather than noise to be removed.
\textit{Second}, $\lambda_{\mathrm{step}}=1.0$ gives the best trade-off ($\lambda_{\mathrm{step}}=0$ corresponds to outcome-only KTO), reported above. Lower step weights underuse the function-level signal, while larger weights begin to regress easier benchmarks without improving the hardest ones.
\textit{Third}, requiring at least two supervised steps outperforms the looser threshold of one, consistent with the redundancy of single-step supervision discussed in Section~\ref{sec:single_step}.

\section{Analysis}
\label{sec:analysis}

\begin{table*}[t]
\centering
\small
\begin{tabular*}{\textwidth}{@{\extracolsep{\fill}}lccccccc}
\toprule
Method
& \multicolumn{2}{c}{HumanEval}
& \multicolumn{2}{c}{MBPP}
& \multicolumn{2}{c}{BigCodeBench}
& LiveCodeBench \\
\cmidrule(lr){2-3}
\cmidrule(lr){4-5}
\cmidrule(lr){6-7}
& Base & Plus & Base & Plus & Full & Hard &  \\
\midrule
\multicolumn{8}{l}{\emph{Conflict-handling policy}} \\
\quad KTO (no step loss)
& \textbf{0.878} & \textbf{0.835} & 0.741 & 0.624 & \textbf{0.367} & 0.115 & 0.116 \\
\quad \stepkto + mask conflicts
& \textbf{0.878} & \textbf{0.835} & 0.741 & 0.624 & \textbf{0.367} & 0.115 & 0.119 \\
\quad \textbf{\stepkto + keep conflicts} (default)
& \textbf{0.878} & \textbf{0.835} & \textbf{0.746} & \textbf{0.630} & \textbf{0.367} & \textbf{0.128} & \textbf{0.127} \\
\midrule
\multicolumn{8}{l}{\emph{Step loss weight $\lambda_{\mathrm{step}}$}} \\
\quad 0.5
& 0.872 & 0.829 & \textbf{0.746} & 0.624 & 0.364 & 0.122 & \textbf{0.127} \\
\quad \textbf{1.0} (default)
& \textbf{0.878} & \textbf{0.835} & \textbf{0.746} & \textbf{0.630} & \textbf{0.367} & \textbf{0.128} & \textbf{0.127} \\
\quad 2.0
& 0.872 & 0.829 & \textbf{0.746} & 0.622 & 0.363 & 0.122 & \textbf{0.127} \\
\midrule
\multicolumn{8}{l}{\emph{Min.\ supervised steps threshold}} \\
\quad 1
& \textbf{0.878} & \textbf{0.835} & 0.741 & 0.627 & 0.363 & 0.115 & 0.116 \\
\quad \textbf{2} (default)
& \textbf{0.878} & \textbf{0.835} & \textbf{0.746} & \textbf{0.630} & \textbf{0.367} & \textbf{0.128} & \textbf{0.127} \\
\bottomrule
\end{tabular*}
\caption{Core ablations on Qwen2.5-Coder-3B-Instruct. We vary conflict handling, the step-loss weight $\lambda_{\mathrm{step}}$, and the minimum number of supervised steps required to activate the step loss.}
\label{tab:ablations}
\end{table*}

\subsection{Why Function-Level Supervision Helps}
\label{sec:why_conflicts}
\label{sec:single_step}

Both DPO and KTO operate at the program level, weighting the policy--reference log-ratio by a single binary outcome label. They therefore cannot distinguish which parts of a generated program contributed to success or failure: when correct and incorrect functions coexist, outcome-level methods reinforce or penalize all components together. \stepkto targets this missing information by assigning labels to individual functions, allowing the model to reinforce locally correct functions and penalize locally incorrect ones within the same program. 
This explains why conflict preservation matters. When $z_i=o$ for all supervised steps, the step loss is directionally aligned with the outcome loss and mainly changes where the gradient is applied. In contrast, when $z_i\neq o$, the step loss provides a direction that outcome-level supervision cannot produce, such as penalizing a specific function even when the full program passes. Such conflicts can arise when the decomposed function interface exposes edge cases not exercised by the original test suite. Masking these conflicts removes this directional disagreement, making the remaining step signal largely redundant with outcome-level supervision.
The same reasoning explains the minimum-supervised-steps threshold. For single-function rows, the step span covers the full answer, so $r_1=r_{\mathrm{out}}$. When $z_1=o$, the step loss adds no information beyond a rescaled outcome loss. Requiring at least two supervised steps filters out these redundant rows.

\subsection{Function-Level Labels Localize Failures}
\label{sec:repair_analysis}

To test whether negative step labels identify functions that contribute to end-to-end failure, we perform a repair intervention on failing candidates whose supervised functions include both positive and negative labels. We replace the locally negative functions with their decomposed reference implementations and evaluate the repaired candidates using the original dataset-provided test suite. As controls, we replace either the same number of locally positive functions or the same number of randomly selected observed functions.

As shown in Table~\ref{tab:repair}, replacing locally negative functions repairs 74.9\% of failing candidates, substantially outperforming both random and locally positive replacement. This suggests that execution-based step labels localize repair-relevant faults, rather than merely correlating with outcome correctness.

\begin{table}[t]
\centering
\small
\begin{tabular*}{\columnwidth}{@{\extracolsep{\fill}}lc}
\toprule
Replacement target & Repair rate \\
\midrule
Local-negative ($z_i = 0$) & \textbf{74.9\%} \\
Random observed & 32.9\% \\
Local-positive ($z_i = 1$) & 1.2\% \\
\bottomrule
\end{tabular*}
\caption{Repair intervention on failing candidates.}
\label{tab:repair}
\end{table}

\subsection{Execution-Based Labels Are Essential}
\label{sec:llm_judge_summary}
A natural question is whether the execution-based unit tests at the core of \stepkto are necessary, or whether an LLM-as-a-judge can provide comparable step-level supervision. We test this by replacing execution-based step labels with judgments from GPT-5.4 mini~\citep{openai2026gpt54mini}, and we retrain Qwen2.5-Coder-3B-Instruct with identical settings.
The LLM shows moderate overall agreement with execution-based labels ($73.2\%$), but this aggregate score hides a strong asymmetry. It agrees with execution on $94.1\%$ of failing functions, but on only $52.2\%$ of passing functions, systematically over-predicting failure. This bias corrupts the positive step labels needed for \stepkto: training with LLM-as-a-judge labels instead of execution-based labels degrades performance, especially on BigCodeBench Hard and LiveCodeBench. Thus, execution-based labels are essential not only because they provide step-level supervision: they also provide reliable step-level supervision. Full agreement statistics and benchmark results across all seven benchmarks are provided in Appendix~\ref{app:llm_judge}.

\section{Conclusion}
\label{sec:conclusion}

We propose \stepkto, a framework for code preference optimization that defines process-supervision steps as module-level functions in decomposed programs. By combining outcome-level KTO with execution-based function labels, \stepkto provides localized feedback that reinforces correct functions while penalizing incorrect ones. Experiments on post-trained code models show consistent gains over outcome-only KTO and DPO, with ablations confirming that preserving local/global label conflicts and using execution-based labels are central to these gains. These results suggest that function-level execution feedback offers a practical path toward process supervision for code generation.

\section{Limitations}
\label{sec:limitations}

\paragraph{Model and data scope.}
\stepkto is most natural when solutions admit a meaningful decomposition into independently testable functions, and is most reliable when candidate solutions are generated on-policy by the target model, which adds engineering cost for
each new model. Our experiments focus on instruction-tuned code models that have already undergone substantial post-training; applying function-level supervision earlier, during SFT or on larger base-model training runs, remains
a promising direction.

\paragraph{Difficulty distribution.}
Our filtering pipeline requires problems to have a clear function-style structure or to admit a meaningful decomposition. As a result, the training data is biased toward problems whose solutions can be decomposed into testable functions, while harder competition problems are less represented. Extending the pipeline to more complex I/O formats and harder problem regimes is a natural next step.

\paragraph{Unit-test quality.}
Step labels depend on automatically generated unit tests, which we validate by execution against decomposed reference solutions but do not formally verify. A mutation-sensitivity audit (Appendix~\ref{app:ut_quality}) shows that the retained unit tests reject 92.6\% of semantically perturbed reference implementations, and our repair intervention (Section~\ref{sec:repair_analysis}) shows that negative step labels identify functions responsible for end-to-end failure. Our LLM-as-a-judge comparison further shows that execution-based labels are substantially more reliable than LLM judgments. Still, stronger automated test-generation or verification methods could further improve label quality.


\section*{Ethical Considerations}

The models used in this paper, Qwen2.5-Coder~\citep{hui2024qwen25coder} and DeepSeek-Coder~\citep{guo2024deepseekcoder}, are licensed for academic research purposes. The training datasets, TACO~\citep{li2023taco} and APPS~\citep{hendrycks2021measuring}, and all evaluation benchmarks~\citep{chen2021evaluatinglargelanguagemodels, liu2023evalplus, austin2021programsynthesislargelanguage, zhuo2025bigcodebench, jain2025livecodebench} are publicly available and distributed for research use.

\section*{Acknowledgments}

This work was supported in part by the National Research Foundation of Korea (NRF) grant funded by the Ministry of Science and ICT (MSIT) (RS-2024-00451435, 20\%; RS-2024-00413957, 20\%), the Institute of Information \& Communications Technology Planning \& Evaluation (IITP) grant funded by the MSIT (RS-2025-02305453, 15\%; RS-2025-02273157, 15\%; RS-2025-25442149, 15\%; RS-2021-II211343, 15\%), the Institute of New Media and Communications (INMAC), the BK21 FOUR program funded by the Ministry of Education, the Artificial Intelligence Graduate School Program (Seoul National University), and the Research Program for Future ICT Pioneers at Seoul National University in 2026.

\newpage
\bibliography{custom}

@inproceedings{lightman2023lets,
  title={Let's Verify Step by Step},
  author={Hunter Lightman and Vineet Kosaraju and Yuri Burda and Harrison Edwards and Bowen Baker and Teddy Lee and Jan Leike and John Schulman and Ilya Sutskever and Karl Cobbe},
  booktitle={The Twelfth International Conference on Learning Representations},
  year={2024},
  url={https://openreview.net/forum?id=v8L0pN6EOi}
}

@misc{luo2024improve,
      title={Improve Mathematical Reasoning in Language Models by Automated Process Supervision}, 
      author={Liangchen Luo and Yinxiao Liu and Rosanne Liu and Samrat Phatale and Meiqi Guo and Harsh Lara and Yunxuan Li and Lei Shu and Yun Zhu and Lei Meng and Jiao Sun and Abhinav Rastogi},
      year={2024},
      eprint={2406.06592},
      archivePrefix={arXiv},
      primaryClass={cs.CL},
      url={https://arxiv.org/abs/2406.06592}, 
}

@inproceedings{lin2025stepkto,
  title = {Step-{KTO}: Optimizing Mathematical Reasoning through Stepwise Binary Feedback},
  author = {Lin, Yen-Ting and
            Jin, Di and
            Xu, Tengyu and
            Wu, Tianhao and
            Sukhbaatar, Sainbayar and
            Zhu, Chen and
            He, Yun and
            Chen, Yun-Nung and
            Weston, Jason E and
            Tian, Yuandong and
            Rahnama, Arash and
            Wang, Sinong and
            Ma, Hao and
            Fang, Han},
  editor = {Valentino, Marco and
            Ferreira, Deborah and
            Thayaparan, Mokanarangan and
            Ranaldi, Leonardo and
            Freitas, Andre},
  booktitle = {Proceedings of The 3rd Workshop on Mathematical Natural Language Processing (MathNLP 2025)},
  month = nov,
  year = {2025},
  address = {Suzhou, China},
  publisher = {Association for Computational Linguistics},
  url = {https://aclanthology.org/2025.mathnlp-main.2/},
  doi = {10.18653/v1/2025.mathnlp-main.2},
  pages = {15--33}
}

@inproceedings{
ethayarajh2024kto,
title={Model Alignment as Prospect Theoretic Optimization},
author={Kawin Ethayarajh and Winnie Xu and Niklas Muennighoff and Dan Jurafsky and Douwe Kiela},
booktitle={Forty-first International Conference on Machine Learning},
year={2024},
url={https://openreview.net/forum?id=iUwHnoENnl}
}

@inproceedings{zhang2025focuseddpo,
  title = {Focused-{DPO}: Enhancing Code Generation Through Focused Preference Optimization on Error-Prone Points},
  author = {Zhang, Kechi and
            Li, Ge and
            Li, Jia and
            Dong, Yihong and
            Li, Jia and
            Jin, Zhi},
  editor = {Che, Wanxiang and
            Nabende, Joyce and
            Shutova, Ekaterina and
            Pilehvar, Mohammad Taher},
  booktitle = {Findings of the Association for Computational Linguistics: ACL 2025},
  month = jul,
  year = {2025},
  address = {Vienna, Austria},
  publisher = {Association for Computational Linguistics},
  url = {https://aclanthology.org/2025.findings-acl.498/},
  doi = {10.18653/v1/2025.findings-acl.498},
  pages = {9578--9591}
}

@misc{dai2024process,
      title={Process Supervision-Guided Policy Optimization for Code Generation}, 
      author={Ning Dai and Zheng Wu and Renjie Zheng and Ziyun Wei and Wenlei Shi and Xing Jin and Guanlin Liu and Chen Dun and Liang Huang and Lin Yan},
      year={2025},
      eprint={2410.17621},
      archivePrefix={arXiv},
      primaryClass={cs.AI},
      url={https://arxiv.org/abs/2410.17621}, 
}

@inproceedings{li2025codeprm,
  title = {{C}ode{PRM}: Execution Feedback-enhanced Process Reward Model for Code Generation},
  author = {Li, Qingyao and
            Dai, Xinyi and
            Li, Xiangyang and
            Zhang, Weinan and
            Wang, Yasheng and
            Tang, Ruiming and
            Yu, Yong},
  editor = {Che, Wanxiang and
            Nabende, Joyce and
            Shutova, Ekaterina and
            Pilehvar, Mohammad Taher},
  booktitle = {Findings of the Association for Computational Linguistics: ACL 2025},
  month = jul,
  year = {2025},
  address = {Vienna, Austria},
  publisher = {Association for Computational Linguistics},
  url = {https://aclanthology.org/2025.findings-acl.428/},
  doi = {10.18653/v1/2025.findings-acl.428},
  pages = {8169--8182}
}

@inproceedings{rafailov2024dpo,
title={Direct Preference Optimization: Your Language Model is Secretly a Reward Model},
author={Rafael Rafailov and Archit Sharma and Eric Mitchell and Christopher D Manning and Stefano Ermon and Chelsea Finn},
booktitle={Thirty-seventh Conference on Neural Information Processing Systems},
year={2023},
url={https://openreview.net/forum?id=HPuSIXJaa9}
}

@misc{guo2024deepseekcoder,
      title={DeepSeek-Coder: When the Large Language Model Meets Programming -- The Rise of Code Intelligence}, 
      author={Daya Guo and Qihao Zhu and Dejian Yang and Zhenda Xie and Kai Dong and Wentao Zhang and Guanting Chen and Xiao Bi and Y. Wu and Y. K. Li and Fuli Luo and Yingfei Xiong and Wenfeng Liang},
      year={2024},
      eprint={2401.14196},
      archivePrefix={arXiv},
      primaryClass={cs.SE},
      url={https://arxiv.org/abs/2401.14196}, 
}

@misc{hui2024qwen25coder,
      title={Qwen2.5-Coder Technical Report}, 
      author={Binyuan Hui and Jian Yang and Zeyu Cui and Jiaxi Yang and Dayiheng Liu and Lei Zhang and Tianyu Liu and Jiajun Zhang and Bowen Yu and Keming Lu and Kai Dang and Yang Fan and Yichang Zhang and An Yang and Rui Men and Fei Huang and Bo Zheng and Yibo Miao and Shanghaoran Quan and Yunlong Feng and Xingzhang Ren and Xuancheng Ren and Jingren Zhou and Junyang Lin},
      year={2024},
      eprint={2409.12186},
      archivePrefix={arXiv},
      primaryClass={cs.CL},
      url={https://arxiv.org/abs/2409.12186}, 
}

@inproceedings{wang2024mathshepherd,
  title = {Math-Shepherd: Verify and Reinforce {LLM}s Step-by-step without Human Annotations},
  author = {Wang, Peiyi and
            Li, Lei and
            Shao, Zhihong and
            Xu, Runxin and
            Dai, Damai and
            Li, Yifei and
            Chen, Deli and
            Wu, Yu and
            Sui, Zhifang},
  editor = {Ku, Lun-Wei and
            Martins, Andre and
            Srikumar, Vivek},
  booktitle = {Proceedings of the 62nd Annual Meeting of the Association for Computational Linguistics (Volume 1: Long Papers)},
  month = aug,
  year = {2024},
  address = {Bangkok, Thailand},
  publisher = {Association for Computational Linguistics},
  url = {https://aclanthology.org/2024.acl-long.510/},
  doi = {10.18653/v1/2024.acl-long.510},
  pages = {9426--9439}
}

@misc{uesato2022solvingmathwordproblems,
      title={Solving math word problems with process- and outcome-based feedback}, 
      author={Jonathan Uesato and Nate Kushman and Ramana Kumar and Francis Song and Noah Siegel and Lisa Wang and Antonia Creswell and Geoffrey Irving and Irina Higgins},
      year={2022},
      eprint={2211.14275},
      archivePrefix={arXiv},
      primaryClass={cs.LG},
      url={https://arxiv.org/abs/2211.14275}, 
}

@inproceedings{zhang2025codedpo,
    title = "{C}ode{DPO}: Aligning Code Models with Self Generated and Verified Source Code",
    author = "Zhang, Kechi  and
      Li, Ge  and
      Dong, Yihong  and
      Xu, Jingjing  and
      Zhang, Jun  and
      Su, Jing  and
      Liu, Yongfei  and
      Jin, Zhi",
    editor = "Che, Wanxiang  and
      Nabende, Joyce  and
      Shutova, Ekaterina  and
      Pilehvar, Mohammad Taher",
    booktitle = "Proceedings of the 63rd Annual Meeting of the Association for Computational Linguistics (Volume 1: Long Papers)",
    month = jul,
    year = "2025",
    address = "Vienna, Austria",
    publisher = "Association for Computational Linguistics",
    url = "https://aclanthology.org/2025.acl-long.771/",
    doi = "10.18653/v1/2025.acl-long.771",
    pages = "15854--15871",
    ISBN = "979-8-89176-251-0"
}

@misc{bavarian2022efficienttraininglanguagemodels,
      title={Efficient Training of Language Models to Fill in the Middle}, 
      author={Mohammad Bavarian and Heewoo Jun and Nikolas Tezak and John Schulman and Christine McLeavey and Jerry Tworek and Mark Chen},
      year={2022},
      eprint={2207.14255},
      archivePrefix={arXiv},
      primaryClass={cs.CL},
      url={https://arxiv.org/abs/2207.14255}, 
}

@inproceedings{
lin2025solvers,
title={Learning to Solve and Verify: A Self-Play Framework for Mutually Improving Code and Test Generation},
author={Zi Lin and Sheng Shen and Jingbo Shang and Jason E Weston and Yixin Nie},
booktitle={NeurIPS 2025 Fourth Workshop on Deep Learning for Code},
year={2025},
url={https://openreview.net/forum?id=j6tMZaPWWF}
}

@inproceedings{ma-etal-2025-dynamic,
    title = "Dynamic Scaling of Unit Tests for Code Reward Modeling",
    author = "Ma, Zeyao  and
      Zhang, Xiaokang  and
      Zhang, Jing  and
      Yu, Jifan  and
      Luo, Sijia  and
      Tang, Jie",
    editor = "Che, Wanxiang  and
      Nabende, Joyce  and
      Shutova, Ekaterina  and
      Pilehvar, Mohammad Taher",
    booktitle = "Proceedings of the 63rd Annual Meeting of the Association for Computational Linguistics (Volume 1: Long Papers)",
    month = jul,
    year = "2025",
    address = "Vienna, Austria",
    publisher = "Association for Computational Linguistics",
    url = "https://aclanthology.org/2025.acl-long.343/",
    doi = "10.18653/v1/2025.acl-long.343",
    pages = "6917--6935",
    ISBN = "979-8-89176-251-0"
}

@inproceedings{
chen2023codet,
title={CodeT:  Code Generation with Generated Tests},
author={Bei Chen and Fengji Zhang and Anh Nguyen and Daoguang Zan and Zeqi Lin and Jian-Guang Lou and Weizhu Chen},
booktitle={The Eleventh International Conference on Learning Representations },
year={2023},
url={https://openreview.net/forum?id=ktrw68Cmu9c}
}

@inproceedings{
hu2022lora,
title={Lo{RA}: Low-Rank Adaptation of Large Language Models},
author={Edward J Hu and Yelong Shen and Phillip Wallis and Zeyuan Allen-Zhu and Yuanzhi Li and Shean Wang and Lu Wang and Weizhu Chen},
booktitle={International Conference on Learning Representations},
year={2022},
url={https://openreview.net/forum?id=nZeVKeeFYf9}
}

@misc{li2023taco,
      title={TACO: Topics in Algorithmic COde generation dataset}, 
      author={Rongao Li and Jie Fu and Bo-Wen Zhang and Tao Huang and Zhihong Sun and Chen Lyu and Guang Liu and Zhi Jin and Ge Li},
      year={2023},
      eprint={2312.14852},
      archivePrefix={arXiv},
      primaryClass={cs.AI},
      url={https://arxiv.org/abs/2312.14852}, 
}

@inproceedings{hendrycks2021measuring,
  title={Measuring Coding Challenge Competence With {APPS}},
  author={Dan Hendrycks and Steven Basart and Saurav Kadavath and Mantas Mazeika and Akul Arora and Ethan Guo and Collin Burns and Samir Puranik and Horace He and Dawn Song and Jacob Steinhardt},
  booktitle={Thirty-fifth Conference on Neural Information Processing Systems Datasets and Benchmarks Track (Round 2)},
  year={2021},
  url={https://openreview.net/forum?id=sD93GOzH3i5}
}

@misc{chen2021evaluatinglargelanguagemodels,
      title={Evaluating Large Language Models Trained on Code}, 
      author={Mark Chen and Jerry Tworek and Heewoo Jun and Qiming Yuan and Henrique Ponde de Oliveira Pinto and Jared Kaplan and Harri Edwards and Yuri Burda and Nicholas Joseph and Greg Brockman and Alex Ray and Raul Puri and Gretchen Krueger and Michael Petrov and Heidy Khlaaf and Girish Sastry and Pamela Mishkin and Brooke Chan and Scott Gray and Nick Ryder and Mikhail Pavlov and Alethea Power and Lukasz Kaiser and Mohammad Bavarian and Clemens Winter and Philippe Tillet and Felipe Petroski Such and Dave Cummings and Matthias Plappert and Fotios Chantzis and Elizabeth Barnes and Ariel Herbert-Voss and William Hebgen Guss and Alex Nichol and Alex Paino and Nikolas Tezak and Jie Tang and Igor Babuschkin and Suchir Balaji and Shantanu Jain and William Saunders and Christopher Hesse and Andrew N. Carr and Jan Leike and Josh Achiam and Vedant Misra and Evan Morikawa and Alec Radford and Matthew Knight and Miles Brundage and Mira Murati and Katie Mayer and Peter Welinder and Bob McGrew and Dario Amodei and Sam McCandlish and Ilya Sutskever and Wojciech Zaremba},
      year={2021},
      eprint={2107.03374},
      archivePrefix={arXiv},
      primaryClass={cs.LG},
      url={https://arxiv.org/abs/2107.03374}, 
}

@misc{austin2021programsynthesislargelanguage,
      title={Program Synthesis with Large Language Models}, 
      author={Jacob Austin and Augustus Odena and Maxwell Nye and Maarten Bosma and Henryk Michalewski and David Dohan and Ellen Jiang and Carrie Cai and Michael Terry and Quoc Le and Charles Sutton},
      year={2021},
      eprint={2108.07732},
      archivePrefix={arXiv},
      primaryClass={cs.PL},
      url={https://arxiv.org/abs/2108.07732}, 
}

@inproceedings{
zhuo2025bigcodebench,
title={BigCodeBench: Benchmarking Code Generation with Diverse Function Calls and Complex Instructions},
author={Terry Yue Zhuo and Vu Minh Chien and Jenny Chim and Han Hu and Wenhao Yu and Ratnadira Widyasari and Imam Nur Bani Yusuf and Haolan Zhan and Junda He and Indraneil Paul and Simon Brunner and Chen Gong and James Hoang and Armel Randy Zebaze and Xiaoheng Hong and Wen-Ding Li and Jean Kaddour and Ming Xu and Zhihan Zhang and Prateek Yadav and Naman Jain and Alex Gu and Zhoujun Cheng and Jiawei Liu and Qian Liu and Zijian Wang and David Lo and Binyuan Hui and Niklas Muennighoff and Daniel Fried and Xiaoning Du and Harm de Vries and Leandro Von Werra},
booktitle={The Thirteenth International Conference on Learning Representations},
year={2025},
url={https://openreview.net/forum?id=YrycTjllL0}
}

@inproceedings{
jain2025livecodebench,
title={LiveCodeBench: Holistic and Contamination Free Evaluation of Large Language Models for Code},
author={Naman Jain and King Han and Alex Gu and Wen-Ding Li and Fanjia Yan and Tianjun Zhang and Sida Wang and Armando Solar-Lezama and Koushik Sen and Ion Stoica},
booktitle={The Thirteenth International Conference on Learning Representations},
year={2025},
url={https://openreview.net/forum?id=chfJJYC3iL}
}

@inproceedings{kwon2023efficient,
author = {Kwon, Woosuk and Li, Zhuohan and Zhuang, Siyuan and Sheng, Ying and Zheng, Lianmin and Yu, Cody Hao and Gonzalez, Joseph and Zhang, Hao and Stoica, Ion},
title = {Efficient Memory Management for Large Language Model Serving with PagedAttention},
year = {2023},
isbn = {9798400702297},
publisher = {Association for Computing Machinery},
address = {New York, NY, USA},
url = {https://doi.org/10.1145/3600006.3613165},
doi = {10.1145/3600006.3613165},
booktitle = {Proceedings of the 29th Symposium on Operating Systems Principles},
pages = {611–626},
numpages = {16},
location = {Koblenz, Germany},
series = {SOSP '23}
}

@inproceedings{
liu2023evalplus,
title={Is Your Code Generated by Chat{GPT} Really Correct? Rigorous Evaluation of Large Language Models for Code Generation},
author={Jiawei Liu and Chunqiu Steven Xia and Yuyao Wang and Lingming Zhang},
booktitle={Thirty-seventh Conference on Neural Information Processing Systems},
year={2023},
url={https://openreview.net/forum?id=1qvx610Cu7}
}

@misc{pan2025modularization,
  title         = {Modularization is Better: Effective Code Generation with Modular Prompting},
  author        = {Pan, Ruwei and Zhang, Hongyu},
  year          = {2025},
  eprint        = {2503.12483},
  archivePrefix = {arXiv},
  primaryClass  = {cs.SE},
  url           = {https://arxiv.org/abs/2503.12483}
}

@misc{openai2026gpt54mini,
  title        = {Introducing {GPT-5.4} mini and nano},
  author       = {{OpenAI}},
  year         = {2026},
  howpublished = {\url{https://openai.com/index/introducing-gpt-5-4-mini-and-nano/}},
  note         = {Accessed: 2026-04-28}
}

@misc{cobbe2021trainingverifierssolvemath,
      title={Training Verifiers to Solve Math Word Problems}, 
      author={Karl Cobbe and Vineet Kosaraju and Mohammad Bavarian and Mark Chen and Heewoo Jun and Lukasz Kaiser and Matthias Plappert and Jerry Tworek and Jacob Hilton and Reiichiro Nakano and Christopher Hesse and John Schulman},
      year={2021},
      eprint={2110.14168},
      archivePrefix={arXiv},
      primaryClass={cs.LG},
      url={https://arxiv.org/abs/2110.14168}, 
}

@inproceedings{10.5555/3600270.3602281,
title={Training language models to follow instructions with human feedback},
author={Long Ouyang and Jeffrey Wu and Xu Jiang and Diogo Almeida and Carroll Wainwright and Pamela Mishkin and Chong Zhang and Sandhini Agarwal and Katarina Slama and Alex Gray and John Schulman and Jacob Hilton and Fraser Kelton and Luke Miller and Maddie Simens and Amanda Askell and Peter Welinder and Paul Christiano and Jan Leike and Ryan Lowe},
booktitle={Advances in Neural Information Processing Systems},
editor={Alice H. Oh and Alekh Agarwal and Danielle Belgrave and Kyunghyun Cho},
year={2022},
url={https://openreview.net/forum?id=TG8KACxEON}
}

@inproceedings{
wei2022finetuned,
title={Finetuned Language Models are Zero-Shot Learners},
author={Jason Wei and Maarten Bosma and Vincent Zhao and Kelvin Guu and Adams Wei Yu and Brian Lester and Nan Du and Andrew M. Dai and Quoc V Le},
booktitle={International Conference on Learning Representations},
year={2022},
url={https://openreview.net/forum?id=gEZrGCozdqR}
}

@inproceedings{ren-etal-2025-alignment,
    title = "Alignment with Fill-In-the-Middle for Enhancing Code Generation",
    author = "Ren, Houxing  and
      Lu, Zimu  and
      Shi, Weikang  and
      Hou, Haotian  and
      Yang, Yunqiao  and
      Wang, Ke  and
      Zhou, Aojun  and
      Pan, Junting  and
      Zhan, Mingjie  and
      Li, Hongsheng",
    editor = "Christodoulopoulos, Christos  and
      Chakraborty, Tanmoy  and
      Rose, Carolyn  and
      Peng, Violet",
    booktitle = "Proceedings of the 2025 Conference on Empirical Methods in Natural Language Processing",
    month = nov,
    year = "2025",
    address = "Suzhou, China",
    publisher = "Association for Computational Linguistics",
    url = "https://aclanthology.org/2025.emnlp-main.419/",
    doi = "10.18653/v1/2025.emnlp-main.419",
    pages = "8304--8320",
    ISBN = "979-8-89176-332-6"
}

@inproceedings{wu-etal-2025-teaching,
    title = "Teaching Your Models to Understand Code via Focal Preference Alignment",
    author = "Wu, Jie  and
      Li, Haoling  and
      Zhang, Xin  and
      Liu, Xiao  and
      Huang, Yangyu  and
      Luo, Jianwen  and
      Zhang, Yizhen  and
      Li, Zuchao  and
      Chu, Ruihang  and
      Yang, Yujiu  and
      Li, Scarlett",
    editor = "Christodoulopoulos, Christos  and
      Chakraborty, Tanmoy  and
      Rose, Carolyn  and
      Peng, Violet",
    booktitle = "Proceedings of the 2025 Conference on Empirical Methods in Natural Language Processing",
    month = nov,
    year = "2025",
    address = "Suzhou, China",
    publisher = "Association for Computational Linguistics",
    url = "https://aclanthology.org/2025.emnlp-main.707/",
    doi = "10.18653/v1/2025.emnlp-main.707",
    pages = "14003--14023",
    ISBN = "979-8-89176-332-6"
}

@inproceedings{li-etal-2025-test,
    title = "{S}*: Test Time Scaling for Code Generation",
    author = "Li, Dacheng  and
      Cao, Shiyi  and
      Cao, Chengkun  and
      Li, Xiuyu  and
      Tan, Shangyin  and
      Keutzer, Kurt  and
      Xing, Jiarong  and
      Gonzalez, Joseph E.  and
      Stoica, Ion",
    editor = "Christodoulopoulos, Christos  and
      Chakraborty, Tanmoy  and
      Rose, Carolyn  and
      Peng, Violet",
    booktitle = "Findings of the Association for Computational Linguistics: EMNLP 2025",
    month = nov,
    year = "2025",
    address = "Suzhou, China",
    publisher = "Association for Computational Linguistics",
    url = "https://aclanthology.org/2025.findings-emnlp.865/",
    doi = "10.18653/v1/2025.findings-emnlp.865",
    pages = "15964--15978",
    ISBN = "979-8-89176-335-7"
}

@inproceedings{
yu2025reasoning,
title={Reasoning Through Execution: Unifying Process and Outcome Rewards for Code Generation},
author={Zhuohao Yu and Weizheng Gu and Yidong Wang and Xingru Jiang and Zhengran Zeng and Jindong Wang and Wei Ye and Shikun Zhang},
booktitle={Forty-second International Conference on Machine Learning},
year={2025},
url={https://openreview.net/forum?id=pLQtovjXiw}
}
\appendix
\newpage

\section{Preference Pair Selection}
\label{app:posneg}

For each problem we generate 8 candidate completions from the target model and select up to one positive and one negative candidate for the training set, based on the outcome label $o$ and the step labels.

\paragraph{Positive candidates.}
Among candidates with $o=1$, we select the one with the cleanest step-level signal: we prioritize candidates whose supervised functions all pass their unit tests, followed by candidates with passing labels and some unknown steps, then candidates containing both passing and failing step labels, and finally candidates without active step supervision. Within each category, ties are broken by preferring candidates with more locally passing functions, fewer locally failing functions, valid parsing, and non-empty outputs.

\paragraph{Negative candidates.}
Among candidates with $o=0$, we select the one with the richest local supervision. We first prioritize candidates that contain both passing and failing supervised functions, since they identify which components remain correct and which are responsible for the failure. If none are available, we fall back to candidates with at least one locally failing function. As a secondary fallback, we use \emph{generated-context labels}, obtained by executing each generated function within the full generated module rather than inserting it into the decomposed reference program. We prioritize candidates with both passing and failing generated-context labels, then candidates with at least one generated-context failure, and finally candidates without informative step signal.

\paragraph{Reference-context and generated-context agreement.}
We compare our default local labels, computed in the reference context, with generated-context labels on selected training rows. They agree on $89.4\%$ of $43{,}458$ comparable supervised steps, consistently across models ($89.2$--$89.6\%$), suggesting that local labels closely align with generated-context behavior while isolating function-level correctness.

\paragraph{Pairing.}
Tasks with only a positive candidate contribute an unpaired positive row, and tasks with only a negative candidate contribute an unpaired negative row; both are usable by KTO but not DPO. Tasks with both contribute one positive and one negative row.

\section{Training Details}
\label{app:training_details}

We fine-tune all target models with LoRA~\citep{hu2022lora} using rank $r=32$, scaling factor $\alpha=64$, dropout $0.05$, maximum sequence length 2048, and global batch size 16. We use AdamW with $(\beta_1,\beta_2)=(0.9,0.95)$ and a cosine learning-rate schedule with 50 warmup steps and minimum ratio $0.1$. We use learning rate $1\times10^{-6}$ for KTO and \stepkto, and $5\times10^{-7}$ for DPO. For DPO and KTO, we set $\beta=0.1$. For \stepkto, we use $\beta_{\mathrm{out}}=\beta_{\mathrm{step}}=0.1$, set $\lambda_{\mathrm{step}}=1.0$, and use unit weights for all KTO value-function coefficients: $\lambda_D=\lambda_U=\lambda_{D,\mathrm{step}}=\lambda_{U,\mathrm{step}}=1.0$. Sensitivity to these coefficients is analyzed in Appendix~\ref{app:lambda_scaling}.

All experiments were run on  2 NVIDIA RTX A5000 GPUs with 24 GB of memory.

\section{Data Construction Cost}
\label{app:construction_cost}
For Qwen2.5-Coder-3B-Instruct, data construction required approximately 78 GPU-hours: 24 GPU-hours for reference decomposition, 43 for function-level unit-test generation, and 11 for sampling $k=8$ candidate completions. Decompositions, skeletons, and tests are generated once and reused across target models; only candidate sampling and CPU-based execution and labeling are model-specific.
\begin{table*}[t]
\centering
\footnotesize
\setlength{\tabcolsep}{3pt}
\begin{tabular}{lccccccc}
\toprule
Method
& \multicolumn{2}{c}{HumanEval}
& \multicolumn{2}{c}{MBPP}
& \multicolumn{2}{c}{BigCodeBench}
& LiveCodeBench \\
\cmidrule(lr){2-3}
\cmidrule(lr){4-5}
\cmidrule(lr){6-7}
& Base & Plus & Base & Plus & Full & Hard & \\
\midrule
KTO
& \msb{0.872}{0.006}
& \ms{0.827}{0.007}
& \msb{0.743}{0.003}
& \msb{0.628}{0.003}
& \ms{0.364}{0.004}
& \ms{0.113}{0.004}
& \ms{0.109}{0.006} \\
\stepkto
& \msb{0.872}{0.006}
& \msb{0.829}{0.006}
& \msb{0.743}{0.004}
& \msb{0.628}{0.003}
& \msb{0.365}{0.003}
& \msb{0.128}{0.000}
& \msb{0.123}{0.006} \\
\bottomrule
\end{tabular}
\caption{Mean pass rate $\pm$ standard deviation across three training seeds. Bold denotes the best mean within each benchmark.}
\label{tab:multiseed}
\end{table*}

\begin{table*}[t]
\centering
\footnotesize
\begin{tabular*}{\textwidth}{@{\extracolsep{\fill}}lccccccc}
\toprule
Method
& \multicolumn{2}{c}{HumanEval}
& \multicolumn{2}{c}{MBPP}
& \multicolumn{2}{c}{BigCodeBench}
& LiveCodeBench \\
\cmidrule(lr){2-3}
\cmidrule(lr){4-5}
\cmidrule(lr){6-7}
& Base & Plus & Base & Plus & Full & Hard & \\
\midrule
\textbf{Qwen2.5-7B-Instruct}
& \pv{0.811} & \pv{0.768} & \pv{0.839} & \pv{0.704} & \pv{0.366} & \pv{0.135} & \pv{0.134} \\
\quad w/ DPO
& 0.817 & 0.768 & 0.844 & 0.706 & 0.362 & 0.149 & \textbf{0.142} \\
\quad w/ Target-DPO
& \pv{0.799} & \pv{0.744} & \pv{0.796} & \pv{0.677} & \pvb{0.376} & \pv{0.108} & \pv{0.138} \\
\quad w/ KTO
& \pv{0.823} & \pv{0.780} & \pvb{0.847} & \pv{0.709} & \pv{0.361} & \pv{0.155} & \pv{0.138} \\
\quad w/ \stepkto
& \pvb{0.829} & \pvb{0.787} & \pv{0.844} & \pvb{0.717} & \pv{0.362} & \pvb{0.162} & \pvb{0.142} \\
\bottomrule
\end{tabular*}
\caption{Pass rates for the general-purpose Qwen2.5-7B-Instruct model.}
\label{tab:qwen7b}
\end{table*}

\section{DPO and Baseline Construction Details}
\label{app:baseline_details}

DPO~\citep{rafailov2024dpo} trains from paired preferences. Given a prompt \(x\), a preferred solution \(y^+\), and a dispreferred solution \(y^-\), its loss is
\[
\mathcal{L}_{\mathrm{DPO}}
=
-\mathbb{E}
\left[
\log \sigma
\left(
\beta
\left(
r_\theta(x,y^+) - r_\theta(x,y^-)
\right)
\right)
\right],
\]
where
\[
r_\theta(x,y)=
\log \frac{\pi_\theta(y\mid x)}
{\pi_{\mathrm{ref}}(y\mid x)}.
\]
Here, \(\pi_\theta\) is the optimized policy, \(\pi_{\mathrm{ref}}\) is the frozen reference policy, \(\sigma(\cdot)\) is the sigmoid function, and \(\beta\) controls the KL regularization strength.

In our experiments, DPO pairs are constructed using the same candidate pool as KTO and \stepkto: a passing candidate is used as \(y^+\) and a failing candidate as \(y^-\). Unlike DPO, KTO and \stepkto operate on individually labeled candidates, allowing them to use unpaired positives or negatives.
KTO samples are created using the same candidate rows and outcome labels as \stepkto, but we discard all function-level step labels, reducing training to outcome-only preference optimization.

We train Target-DPO~\citep{wu-etal-2025-teaching} on its dataset of 59{,}000 preference pairs. For comparability, we match our DPO LoRA setup: $r=32$, $\alpha=64$, dropout $0.05$, global batch size 16, and learning rate $5\times10^{-7}$.

\section{Multi-Seed Robustness}
\label{app:multiseed}

To assess seed sensitivity, we train Qwen2.5-Coder-3B-Instruct with both KTO and \stepkto across three random seeds, holding all other settings fixed. Table~\ref{tab:multiseed} reports the resulting mean pass rate and standard deviation.

\stepkto exceeds seed-matched KTO in all three runs on both BigCodeBench Hard
and LiveCodeBench. The corresponding mean gains are 0.015 and 0.014,
respectively; on the other five benchmarks, the difference between the method
means is at most 0.002.

\section{General-Purpose Model}
\label{app:qwen7b}
To test whether this pattern persists for a general-purpose
instruction-tuned model, we additionally train Qwen2.5-7B-Instruct using on-policy candidates together with the validated ground-truth anchors used in our Qwen experiments. Qwen2.5-7B-Instruct starts from a stronger base than our other target models on MBPP, BigCodeBench, and LiveCodeBench.
\stepkto exceeds KTO on six of seven benchmarks, with the largest gains on BigCodeBench Hard and LiveCodeBench; see Table~\ref{tab:qwen7b} for full results.

\section{Dataset Statistics}
\label{app:dataset_stats}

This section reports difficulty distributions across the three stages of our data construction pipeline---the raw TACO and APPS training splits, the problems retained after ground-truth decomposition validation, and the final per-source training rows used by each target model---together with quality statistics for the function-level unit tests generated in Stage 2.

\paragraph{Raw datasets.} Table~\ref{tab:raw_dataset_stats} reports the difficulty composition of the raw TACO and APPS training splits as released. Both datasets contain difficulty annotations covering a wide range from introductory exercises to competition-level problems. Our pipeline retains only problems with an explicit function-style entry point---identified by a non-empty \texttt{fn\_name} field in the dataset's annotations---and discards problems with other interfaces, which typically contain monolithic algorithmic solutions resistant to function-level decomposition. As a result, higher-difficulty buckets contribute few or no problems to our training set: in TACO, only \texttt{easy}, \texttt{medium}, and \texttt{medium\_hard} problems contribute, while \texttt{hard}, \texttt{very\_hard}, and \texttt{unknown\_difficulty} problems are filtered out entirely. In APPS, only \texttt{introductory} and \texttt{interview} problems remain; \texttt{competition} problems are excluded.

\begin{table}[t]
\centering
\small
\begin{tabular*}{\columnwidth}{@{\extracolsep{\fill}}llrr}
\toprule
Source & Difficulty & \# Problems & \% \\
\midrule
\multirow{6}{*}{TACO}
  & easy                & 8{,}904  & 35.0 \\
  & medium              & 3{,}244  & 12.7 \\
  & medium\_hard        & 2{,}745  & 10.8 \\
  & hard                & 3{,}162  & 12.4 \\
  & very\_hard          & 2{,}374  &  9.3 \\
  & unknown\_difficulty & 5{,}014  & 19.7 \\
\cmidrule(lr){2-4}
  & Total               & 25{,}443 & 100.0 \\
\midrule
\multirow{2}{*}{APPS}
  & introductory        & 2{,}353 & 84.0 \\
  & interview           &   450   & 16.0 \\
\cmidrule(lr){2-4}
  & Total               & 2{,}803 & 100.0 \\
\bottomrule
\end{tabular*}
\caption{Difficulty distribution of the raw TACO and APPS training splits before any filtering.}
\label{tab:raw_dataset_stats}
\end{table}

\paragraph{Ground-truth decomposition validation.}
Table~\ref{tab:gt_validation} reports per-difficulty pass rates after Stage 1, where ground-truth solutions are decomposed with Qwen2.5-Coder-32B-Instruct and validated by execution against the original test suite. In the table, \emph{Total} denotes the number of problems entering Stage 1 with a clear function-style entry point, and \emph{Validated} denotes the number whose decomposed solution still passes the original tests. Only validated problems proceed to candidate generation. Validation is much more permissive on APPS ($88.9\%$) than on TACO ($43.9\%$). The TACO drop is concentrated in the harder buckets: \texttt{easy} validates at $52.1\%$, \texttt{medium\_hard} at $37.1\%$, and \texttt{medium} at only $18.8\%$, reflecting the difficulty of producing semantically equivalent decompositions for problems with complex global state.

\begin{table}[t]
\centering
\small
\begin{tabular}{llrrr}
\toprule
Source & Difficulty & Total & Validated & Pass rate \\
\midrule
\multirow{3}{*}{TACO}
  & easy         & 4{,}112 & 2{,}141 & 52.1\% \\
  & medium       & 1{,}200 &   225   & 18.8\% \\
  & medium\_hard &   520   &   193   & 37.1\% \\
\cmidrule(lr){2-5}
  & Total        & 5{,}832 & 2{,}559 & 43.9\% \\
\midrule
\multirow{2}{*}{APPS}
  & introductory & 2{,}325 & 2{,}069 & 89.0\% \\
  & interview    &   442   &   390   & 88.2\% \\
\cmidrule(lr){2-5}
  & Total        & 2{,}767 & 2{,}459 & 88.9\% \\
\bottomrule
\end{tabular}
\caption{Ground-truth decomposition validation rates by difficulty.}
\label{tab:gt_validation}
\end{table}

\begin{table*}[t]
\centering
\small
\begin{tabular*}{\textwidth}{@{\extracolsep{\fill}}ll r rr r}
\toprule
& & \multicolumn{3}{c}{Qwen2.5-Coder family} 
  & \multicolumn{1}{c}{DeepSeek-Coder} \\
\cmidrule(lr){3-5}
\cmidrule(lr){6-6}
Source & Difficulty
& Ground truth anchors
& 1.5B gen.
& 3B gen.
& 6.7B gen. \\
\midrule
\multirow{3}{*}{TACO}
  & easy          & 2{,}141 & 2{,}671 & 2{,}695 & 2{,}748 \\
  & medium        &    225  &    271  &    278  &    301 \\
  & medium\_hard  &    193  &    245  &    237  &    260 \\
\midrule
\multirow{2}{*}{APPS}
  & introductory  & 2{,}069 & 2{,}643 & 2{,}657 & 2{,}661 \\
  & interview     &    390  &    549  &    555  &    603 \\
\midrule
\multicolumn{2}{l}{\textbf{Total rows}}
& \textbf{5{,}018}
& \textbf{6{,}379}
& \textbf{6{,}422}
& \textbf{6{,}573} \\
\bottomrule
\end{tabular*}
\caption{Difficulty distribution of the final training sets for each target model. Qwen2.5-Coder-1.5B-Instruct and Qwen2.5-Coder-3B-Instruct use the same validated decomposed reference solutions as positive anchors, but differ in their on-policy generated candidates. DeepSeek-Coder-6.7B-Instruct uses only on-policy generated candidates.}
\label{tab:final_dataset_stats}
\end{table*}

\paragraph{Final training sets.} Table~\ref{tab:final_dataset_stats} reports the difficulty distribution of the rows actually used to train each target model. For Qwen2.5-Coder-3B-Instruct (same-family), we retain all 5{,}018 validated ground-truth decompositions as additional positive anchors and add 6{,}422 generated rows produced by the target model itself. For DeepSeek-Coder-6.7B-Instruct (cross-family), ground-truth rows are discarded (Appendix~\ref{app:gt_ablation}) and only on-policy generated candidates are retained.


The final training sets are skewed toward easier problems, reflecting both the natural distribution of decomposition-friendly problems and the per-difficulty validation rates from Table~\ref{tab:gt_validation}. We discuss the implications in the \emph{Difficulty distribution} paragraph of Section~\ref{sec:limitations}.

\section{Unit Test Generation Quality}
\label{app:ut_quality}
\paragraph{Generation and validation protocol.}
Function-level unit tests are produced by Qwen2.5-Coder-32B-Instruct (Section~\ref{sec:pipeline}, Stage 2) and validated by execution against the decomposed reference implementation. The generator outputs structured JSON test cases with named inputs, which are converted into a \texttt{unittest} class that calls the target function on each input and compares the result to the reference output. Tests are validated by execution: only test classes that parse, execute, and pass against the ground-truth code are retained. Functions for which no valid test remains receive a null step label ($z_i = \varnothing$) and fall back to outcome-only KTO supervision during training. Table~\ref{tab:ut_quality} summarizes unit-test generation quality, and Figure~\ref{fig:ut_coverage} shows the full validated-test-availability distribution.

\begin{table}[t]
\centering
\small
\begin{tabular*}{\columnwidth}{@{\extracolsep{\fill}}lrr}
\toprule
& TACO & APPS \\
\midrule
\multicolumn{3}{l}{\textit{Decomposition scale}} \\
\quad Validated problems & 2{,}559 & 2{,}459 \\
\quad Total functions & 7{,}199 & 6{,}926 \\
\quad Testable functions & 7{,}144 & 6{,}920 \\
\midrule
\multicolumn{3}{l}{\textit{Unit-test generation}} \\
\quad First-pass parse success & 98.8\% & 99.6\% \\
\quad Testable functions w/ $\geq 1$ valid test & 82.0\% & 84.5\% \\
\quad Mean valid tests / covered function & 5.46 & 5.59 \\
\quad Mean per-problem test availability & 83.4\% & 85.4\% \\
\midrule
\multicolumn{3}{l}{\textit{Excluded functions}} \\
\quad Skipped functions (I/O glue) & 55 & 6 \\
\bottomrule
\end{tabular*}
\caption{Function-level unit test generation quality. Validated-test availability is over testable functions.}
\label{tab:ut_quality}
\end{table}

\begin{table*}[t]
\centering
\small
\begin{tabular*}{\textwidth}{@{\extracolsep{\fill}}lccccccc}
\toprule
Configuration
& \multicolumn{2}{c}{HumanEval}
& \multicolumn{2}{c}{MBPP}
& \multicolumn{2}{c}{BigCodeBench}
& LiveCodeBench \\
\cmidrule(lr){2-3}
\cmidrule(lr){4-5}
\cmidrule(lr){6-7}
& Base & Plus & Base & Plus & Full & Hard & \\
\midrule
\textbf{$\lambda_U = \lambda_{U,\mathrm{step}} = 1.0$} (default)
& \textbf{0.878} & \textbf{0.835} & \textbf{0.746} & \textbf{0.630} & \textbf{0.367} & \textbf{0.128} & \textbf{0.127} \\
$\lambda_{U,\mathrm{step}} = 2.5$
& \textbf{0.878} & \textbf{0.835} & \textbf{0.746} & 0.624 & 0.365 & 0.122 & \textbf{0.127} \\
$\lambda_U = 2.5$
& \textbf{0.878} & \textbf{0.835} & \textbf{0.746} & 0.624 & 0.365 & 0.122 & 0.049 \\
\bottomrule
\end{tabular*}
\caption{Effect of rebalancing outcome- and step-level $\lambda_U$ on Qwen2.5-Coder-3B-Instruct.}
\label{tab:lambda_ablation}
\end{table*}

\begin{table*}[t]
\centering
\small
\begin{tabular*}{\textwidth}{@{\extracolsep{\fill}}lccccccc}
\toprule
Method
& \multicolumn{2}{c}{HumanEval}
& \multicolumn{2}{c}{MBPP}
& \multicolumn{2}{c}{BigCodeBench}
& LiveCodeBench \\
\cmidrule(lr){2-3}
\cmidrule(lr){4-5}
\cmidrule(lr){6-7}
& Base & Plus & Base & Plus & Full & Hard & \\
\midrule
KTO only
& \textbf{0.878} & \textbf{0.835} & 0.741 & 0.624 & \textbf{0.367} & 0.115 & 0.116 \\
Step only
& 0.866 & 0.823 & 0.741 & \textbf{0.630} & 0.360 & 0.100 & 0.112 \\
\textbf{\stepkto{}} (default)
& \textbf{0.878} & \textbf{0.835} & \textbf{0.746} & \textbf{0.630} & \textbf{0.367} & \textbf{0.128} & \textbf{0.127} \\
\bottomrule
\end{tabular*}
\caption{Step-only objective ablation on Qwen2.5-Coder-3B-Instruct.}
\label{tab:step_only_ablation}
\end{table*}

As shown in Table~\ref{tab:ut_quality}, first-pass parsing succeeds in over $98\%$ of calls, and covered functions have around $5.5$ valid tests on average, providing multiple independent assertions per supervised step. Overall, $83.2\%$ of testable decomposed functions retain at least one
validated test, while mean per-problem validated-test availability is $84.4\%$; the latter is visualized in Figure~\ref{fig:ut_coverage}.

\paragraph{Mutation-sensitivity audit.}
As an additional sanity check, we evaluate whether the generated function-level tests detect controlled perturbations rather than merely executing successfully on the reference implementation. We audit all 11{,}705 functions from validated decomposed programs that retained at least one valid generated unit test across TACO and APPS, and apply syntax-preserving mutations to the reference function, including comparison flips, Boolean-operator flips, arithmetic-operator changes, constant perturbations, and default-return replacements. Across 27{,}958 valid mutants, the generated tests reject 25{,}880 mutants, yielding a mutation kill rate of 92.6\%. This suggests that the retained tests reliably detect local behavioral changes rather than only validating executability on the reference solution.

\section{Representative Function-Level Label Cases}
\label{app:label_cases}

\paragraph{Incomplete local test.}
For a Pair of Shoes task (TACO 11007), the candidate checks whether the concatenated
left and right-shoe sizes are unique, rather than comparing the two size multisets. Because the retained test for \texttt{pair\_of\_shoes} covers only the empty input, both local labels are positive although the full program is incorrect. This illustrates why \stepkto retains the outcome-level term.
\paragraph{Useful local/global conflict.}
In a Task Scheduler problem (APPS 171), the candidate passes the dataset-provided test suite, but its decomposed step \texttt{calculate\_min\_intervals} omits \texttt{len(tasks)} from the maximum, understating the required number of intervals on inputs where the task count exceeds the frequency-based bound. Its negative local label
exposes a hidden defect that outcome-only KTO cannot represent.

\section{Repair Intervention Details}
\label{app:repair_analysis}
Beyond the aggregate repair rates reported in Section~\ref{sec:repair_analysis}, we also examine how the intervention behaves across failure types. The analysis is run on originally failing Qwen2.5-Coder-3B-Instruct selected training candidates whose supervised functions include both positive and negative step labels. For each candidate, we replace local-negative functions with their decomposed reference implementations and rerun the dataset-provided tests. As controls, we replace either the same number of local-positive functions or the same number of randomly selected observed functions.

The repair effect is strongest for wrong-answer failures: replacing local-negative functions repairs 78.6\% of such candidates on the combined APPS and TACO subset. This suggests that local-negative labels are especially effective at identifying semantic errors, rather than merely capturing parsing or execution artifacts.

\section{Sensitivity to KTO Value-Function Weights}
\label{app:lambda_scaling}

The default configuration uses unit weights for all KTO value-function coefficients
($\lambda_D = \lambda_U = \lambda_{D,\mathrm{step}} = \lambda_{U,\mathrm{step}} = 1.0$).
The KTO authors recommend rebalancing $\lambda_U$ when the positive/negative sample ratio is skewed; our Qwen2.5-Coder-3B-Instruct training set has a step-level positive/negative ratio of approximately $2.5{:}1$, motivating a check on whether upweighting undesirable samples improves performance. When one coefficient is varied, all other value-function coefficients remain fixed at $1.0$. Table~\ref{tab:lambda_ablation} reports two rebalancing runs against the default.

Both deviations from uniform weights regress BigCodeBench Hard. The outcome-level $\lambda_U = 2.5$ run additionally collapses LiveCodeBench performance, suggesting that outcome-level rebalancing is unstable in this setting. One possible explanation is that the observed positive/negative ratio reflects the natural pass distribution of a strong instruction-tuned 3B model rather than simple dataset imbalance; reweighting failures too aggressively may over-allocate gradient budget toward avoiding failures at the cost of reinforcing successful behavior, hurting generalization to harder benchmarks.

\section{Additional Objective Ablations}
\label{app:objective_ablations}

Our default objective combines outcome-level and function-level supervision:
\[
\mathcal{L}_{\text{\stepktomath}}
=
\mathcal{L}_{\mathrm{out}}
+
\lambda_{\mathrm{step}}\mathcal{L}_{\mathrm{step}}.
\]
To better understand the role of the outcome-level term, we evaluate a step-only variant on Qwen2.5-Coder-3B-Instruct. This variant removes the outcome-level KTO loss and optimizes only the stepwise loss, allowing us to assess how much signal is provided by local execution feedback alone.

As shown in Table~\ref{tab:step_only_ablation}, the step-only variant retains useful signal from function-level execution feedback, matching KTO on MBPP and matching the full \stepkto objective on MBPP+. However, removing the outcome-level KTO term weakens performance on HumanEval, BigCodeBench, and LiveCodeBench. The \stepkto objective performs best overall, with the clearest advantage on BigCodeBench Hard. These results support our formulation of \stepkto as a joint objective: the step loss provides localized credit assignment, while the outcome-level term preserves the end-to-end correctness signal needed for robust program-level performance.

\section{Off-Policy Training Data in Cross-Family Training}
\label{app:gt_ablation}

To test whether on-policy candidate generation is necessary, we train DeepSeek-Coder-6.7B-Instruct on candidates generated by Qwen2.5-Coder-3B-Instruct on the same set of decomposed problems used in our main DeepSeek run. Both training sets share the same decomposed function skeletons and unit tests, generated by Qwen2.5-Coder-32B-Instruct; the only difference is the model that produced the candidate completions. The Qwen-generated candidates are fully off-policy with respect to the DeepSeek reference model.

\begin{table*}[t]
\centering
\small
\setlength{\tabcolsep}{4.92pt}
\begin{tabular}{llccccccc}
\toprule
Training data & Method
& \multicolumn{2}{c}{HumanEval}
& \multicolumn{2}{c}{MBPP}
& \multicolumn{2}{c}{BigCodeBench}
& LiveCodeBench \\
\cmidrule(lr){3-4}
\cmidrule(lr){5-6}
\cmidrule(lr){7-8}
& & Base & Plus & Base & Plus & Full & Hard & \\
\midrule
On-policy (default) & KTO         & 0.793 & 0.726 & 0.749 & 0.640 & 0.347 & 0.122 & 0.127 \\
On-policy (default) & \textbf{\stepkto}    & 0.799 & 0.732 & \textbf{0.754} & 0.648 & \textbf{0.347} & \textbf{0.128} & \textbf{0.131} \\
\midrule
Off-policy (Qwen 3B-generated) & KTO         & \textbf{0.805} & \textbf{0.735} & 0.749 & 0.640 & 0.345 & 0.115 & 0.112 \\
Off-policy (Qwen 3B-generated) & \stepkto    & \textbf{0.805} & \textbf{0.735} & 0.751 & \textbf{0.648} & \textbf{0.347} & 0.122 & 0.127 \\
\bottomrule
\end{tabular}
\caption{Effect of training with off-policy data on DeepSeek-Coder-6.7B-Instruct.}
\label{tab:off_policy_data}
\end{table*}

As shown in Table~\ref{tab:off_policy_data}, off-policy training degrades both methods on the hardest benchmark (BigCodeBench Hard: KTO 0.122 to 0.115, \stepkto 0.128 to 0.122). The effect on LiveCodeBench is asymmetric: KTO regresses substantially (0.127 to 0.112) while \stepkto regresses marginally (0.131 to 0.127), suggesting that function-level supervision is more robust to policy mismatch than outcome-only supervision. Easier benchmarks (HumanEval, HumanEval+) show small improvements under off-policy training, likely reflecting the broader problem distribution covered by the Qwen model's candidate pool.

We attribute the BigCodeBench Hard regression to noisier log-ratio rewards: when the candidate distribution diverges from the reference model's, the ratio \(\log \pi_\theta / \pi_{\text{ref}}\) is poorly calibrated, weakening the KTO objective. The function-level signal partially compensates because it operates on shorter token spans, where calibration noise has less cumulative effect. We therefore use the on-policy configuration as our default setting.

\section{LLM-as-a-Judge vs.\ Execution-Based Step Labels}
\label{app:llm_judge}

To evaluate execution-based step labeling against a common alternative, we compare \stepkto's unit-test-based function labels with annotations from GPT-5.4 mini, applied to the same candidate functions in the Qwen2.5-Coder-3B-Instruct training pipeline. The comparison has two components: (i) label-level agreement statistics on the full candidate pool, and (ii) a downstream training experiment in which execution-based labels are replaced by LLM-as-a-judge labels in the actual \stepkto training set.

\paragraph{Setup.}
For agreement analysis, we sample $N{=}88{,}469$ supervised functions from the Qwen2.5-Coder-3B-Instruct candidate pool, stratified by the $(o, z)$ outcome--step pair, and ask GPT-5.4 mini to predict whether each function would pass its associated unit tests. The judge sees the function source code and the unit tests, and returns a binary label with a one-sentence rationale; the system prompt is reproduced in Appendix~\ref{app:prompts}. For the downstream experiment, we apply the same procedure only to the supervised functions in the final training rows, replace execution-based labels with LLM-as-a-judge labels where available, and retrain with the same settings as the main result.

\begin{figure}[!t]
    \centering
    \captionsetup{font=normalsize,skip=3pt}
    \captionsetup[subfigure]{font=small,skip=3pt}

    \begin{subfigure}{\columnwidth}
        \centering
        \includegraphics[width=0.79\columnwidth]{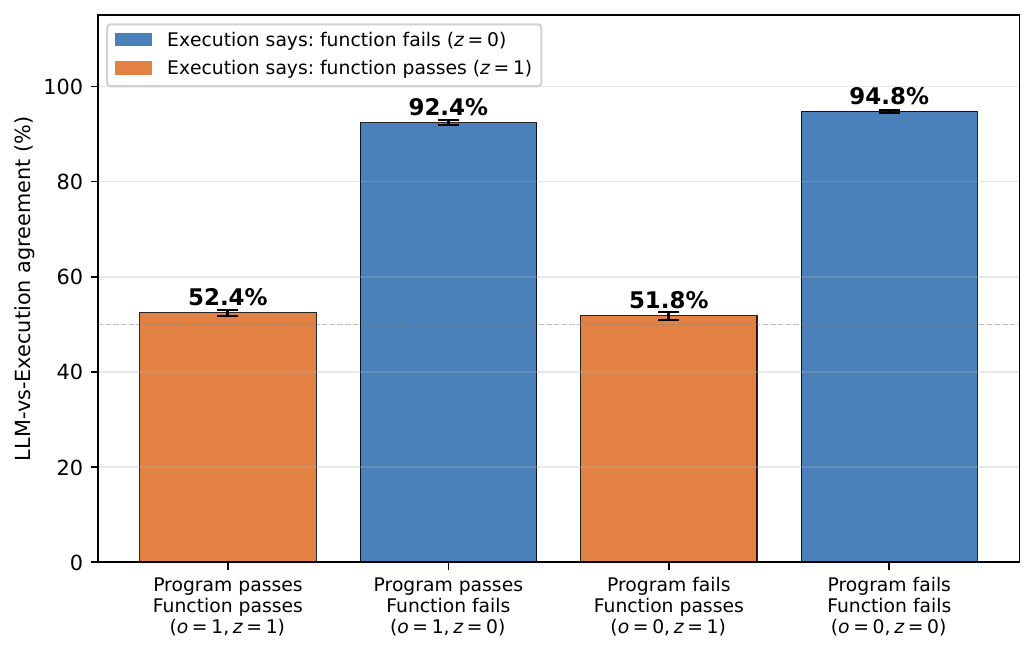}
        \caption{Agreement by $(o,z)$ stratum.}
        \label{fig:agreement_by_stratum}
    \end{subfigure}

    \vspace{0.5em}

    \begin{subfigure}{\columnwidth}
        \centering
        \includegraphics[width=0.79\columnwidth]{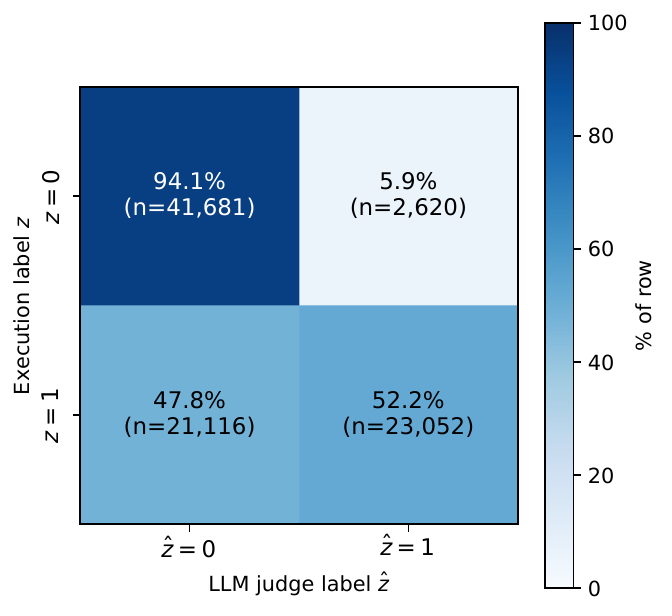}
        \caption{Row-normalized confusion matrix.}
        \label{fig:llm_confusion}
    \end{subfigure}
    
    \vspace{0.13em}
    
    \caption{LLM-as-a-judge agreement with execution-based labels is high for failing functions ($z{=}0$) but much lower for passing functions ($z{=}1$), revealing a bias toward predicting failure.}
    \label{fig:llm_judge_agreement}
\end{figure}

\begin{table*}[!t]
\centering
\small
\setlength{\tabcolsep}{4.30pt}
\begin{tabular}{lccccccc}
\toprule
Step labels
& \multicolumn{2}{c}{HumanEval}
& \multicolumn{2}{c}{MBPP}
& \multicolumn{2}{c}{BigCodeBench}
& LiveCodeBench \\
\cmidrule(lr){2-3}
\cmidrule(lr){4-5}
\cmidrule(lr){6-7}
& Base & Plus & Base & Plus & Full & Hard & \\
\midrule
KTO baseline
& \textbf{0.878} & \textbf{0.835} & 0.741 & 0.624 & \textbf{0.367} & 0.115 & 0.116 \\
\textbf{\stepkto{} + execution} (default)
& \textbf{0.878} & \textbf{0.835} & \textbf{0.746} & 0.630 & \textbf{0.367} & \textbf{0.128} & \textbf{0.127} \\
\stepkto{} + LLM-as-a-judge (GPT-5.4 mini)
& 0.860 & 0.817 & \textbf{0.746} & \textbf{0.635} & 0.354 & 0.095 & 0.112 \\
\quad \emph{Relative change vs.\ execution}
& \emph{-2.1\%} & \emph{-2.2\%} & \emph{+0.0\%} & \emph{+0.8\%} & \emph{-3.5\%} & \emph{-25.8\%} & \emph{-11.8\%} \\
\bottomrule
\end{tabular}
\caption{Effect of replacing execution-based step labels with GPT-5.4 mini judgments on Qwen2.5-Coder-3B-Instruct.}
\label{tab:llm_judge_training}
\end{table*}

\paragraph{Agreement is asymmetric.}
Figure~\ref{fig:agreement_by_stratum} reports agreement by $(o,z)$ stratum, and Figure~\ref{fig:llm_confusion} shows the row-normalized confusion matrix. Overall agreement is $73.2\%$, but this aggregate score hides a strong asymmetry. The LLM agrees with execution on $94.1\%$ of functions labeled as failing by execution ($z{=}0$), but on only $52.2\%$ of functions labeled as passing by execution ($z{=}1$). At the stratum level, agreement is high on both failing-function strata ($o{=}1,z{=}0$: $92.4\%$; $o{=}0,z{=}0$: $94.8\%$), but only around $52\%$ on both passing-function strata.

This pattern indicates that the LLM judge has a strong false-negative bias: it often flags true failures, but also frequently hallucinates issues in correct implementations. This preserves many negative conflict labels while corrupting the much larger pool of clean positive step labels.

\paragraph{Downstream impact.}
Table~\ref{tab:llm_judge_training} compares \stepkto trained with execution-based labels versus LLM-as-a-judge labels on Qwen2.5-Coder-3B-Instruct. Replacing execution labels with GPT-5.4 mini judgments substantially degrades the hardest benchmarks, with relative drops of $25.8\%$ on BigCodeBench Hard and $11.8\%$ on LiveCodeBench. Notably, BigCodeBench Hard falls below the outcome-only KTO baseline, confirming that moderate label-level agreement can still hide systematic bias that harms downstream training.

\paragraph{Discussion.}
Moderate aggregate agreement does not translate into useful supervision because the LLM judge's errors are highly asymmetric: its false-negative bias corrupts passing step labels, which \stepkto needs to stabilize function-level training. Within the operating regime of our framework, execution-based labels are not merely a convenient choice but a necessary one.

\section{Prompts}
\label{app:prompts}

Here, we outline the key prompts used in our \stepkto data construction pipeline. Figure~\ref{fig:prompt_decomp} shows the prompts used to decompose reference solutions into multi-function programs. Figure~\ref{fig:prompt_unit_tests_user} gives the prompt used for generating function-level unit tests, and Figure~\ref{fig:prompt_candidate_user} shows the prompt used to sample on-policy candidate completions from function skeletons. Finally, Figure~\ref{fig:prompt_judge} gives the prompt used by the LLM-as-a-judge in our labeling-method comparison (Appendix~\ref{app:llm_judge}).

\begin{figure*}[t]

\begin{promptbox}{Prompt for Function Decomposition}
\textcolor{blue!60!black}{\textbf{SYSTEM PROMPT}}\\[1mm]
You are an expert Python refactoring assistant.\\
Your task is to decompose a Python solution into semantically meaningful top-level helper functions while preserving exact behavior and the exact required interface.\\
When the solution contains distinct logical stages, decompose it into multiple helper functions rather than leaving everything inside one large function.\\
Do not invent trivial helpers or change the algorithm unnecessarily. Follow the requested XML schema exactly. Return ONLY valid XML. No explanations.

\par\medskip
\hrule
\medskip

\textcolor{blue!60!black}{\textbf{USER PROMPT}}\\[1mm]
Refactor the following Python solution into structured XML with multiple <function> blocks.\\[2mm]
Goal:\\
- Decompose into helper functions where useful.\\
- Preserve behavior exactly.\\
- Keep the required entrypoint unchanged (same name and signature).\\[2mm]
Required entrypoint (must appear EXACTLY in your code):\\
\texttt{\{required\_entrypoint\_line\}}\\[2mm]
Rules:\\
1. Do NOT introduce class Solution (this is a plain function problem).\\
2. Do NOT read from stdin or print output.\\
3. Do NOT include any top-level execution (no main(), no \_\_starting\_point(), no if \_\_name\_\_ == "\_\_main\_\_": ...).\\
4. Put helper functions ABOVE the required entrypoint.\\
5. Each function definition must be in its own <function> block.\\
6. The <code> section should contain exactly ONE def statement (or one class).\\
7. Copy all original imports at the very top of the FIRST <code> block.\\
8. Do NOT add unit tests yet: keep <tests> as placeholder comments.\\
9. Do NOT define nested functions inside another function or method. Every helper must be top-level and placed in its own <function> block.\\
10. When all <code> blocks are concatenated in order, the result must be a valid standalone Python solution with exactly the same behavior as the original.\\
11. When the solution contains distinct logical stages, decompose it into multiple semantically meaningful helper functions.\\
12. Avoid returning a single large function unless decomposition is genuinely unnecessary.\\
13. Do not invent trivial helpers just to increase the number of functions.\\[2mm]
XML format:\\
<function name="..."> <docstring>...</docstring> <tests> \{tests\_placeholder\} </tests> <code> ... code here ... </code> </function>\\[2mm]
Problem: \{problem\_description\}\\[1mm]
Original solution: \{ground\_truth\_solution\}\\[2mm]
Refactored XML format (ONLY XML):
\end{promptbox}

\caption{System and user prompts used to decompose reference solutions into behavior-preserving multi-function programs.}
\label{fig:prompt_decomp}

\end{figure*}

\begin{figure*}[t]
\begin{promptbox}{Prompt for Unit Test Generation}
Output must be a JSON object with this exact schema:\\
\{\ "cases": [ \{"name": "short\_name", "args": [...], "kwargs": \{...\}\}, ... ] \}\\[2mm]
Rules:\\
- args must be a JSON array; kwargs must be a JSON object (use \{\} if none).\\
- Use only JSON-serializable values: null, true/false, numbers, strings, lists, objects.\\
- Keep inputs physically SMALL (e.g., arrays under 5 items, integers between -50 and 50) to avoid execution timeouts.\\
- The test suite MUST include at least one or two ADVERSARIAL EDGE CASES (e.g., empty lists [], empty strings "", zero 0, or negative numbers -1).\\
- Do NOT make every case an edge case. Provide a balanced mix of typical and boundary inputs.\\
- Do NOT include expected outputs.\\
- Provide 4 to 6 diverse cases.
\end{promptbox}
\caption{User prompt schema and constraints for generating function-level unit tests. The model returns only test inputs; expected outputs are derived by executing the reference implementation.}
\label{fig:prompt_unit_tests_user}
\end{figure*}

\begin{figure*}[t]
\begin{promptbox}{Prompt for Candidate Generation}
Solve the following programming problem.\\[2mm]
You MUST implement the provided skeleton by replacing each pass with a correct implementation.\\[2mm]
Rules:\\
- Keep all function/class names and signatures unchanged.\\
- Do not remove any definitions.\\
- Do not add new functions, methods, or classes.\\
- Only fill the bodies of the provided definitions.\\
- Return a single complete Python module.\\[2mm]
\#\#\# problem\\
\{question\}\\[2mm]
\#\#\# skeleton\\
\{skeleton\_code\}
\end{promptbox}
\caption{User prompt for candidate generation. The target model fills the function skeleton (with \texttt{pass} as each function body) constructed in Stage 3 of our pipeline.}
\label{fig:prompt_candidate_user}
\end{figure*}

\begin{figure*}[t]
\begin{promptbox}{Prompt for LLM-as-a-Judge}
You are an expert Python programmer acting as a code reviewer. You will be given a Python function and a set of unit tests for that function. Your job is to decide whether the function implementation would pass all the provided unit tests when executed.\\[2mm]
Respond in strict JSON of the form:\\
\{"label": 0 or 1, "reason": "<one short sentence>"\}\\[2mm]
- label = 1 means you believe the function would pass all provided tests.\\
- label = 0 means you believe at least one test would fail.\\[2mm]
Do not include any other text outside of the JSON object.
\end{promptbox}
\caption{System prompt for the GPT-5.4 mini judge in the LLM-as-a-judge experiment (Appendix~\ref{app:llm_judge}). The judge sees the function's source code and its generated unit tests, and predicts whether the function passes them.}
\label{fig:prompt_judge}
\end{figure*}

\end{document}